\documentclass[lettersize,journal]{IEEEtran}
\usepackage{amsmath,amsfonts}
\usepackage{array}
\usepackage[caption=false,font=normalsize,labelfont=sf,textfont=sf]{subfig}
\usepackage{textcomp}
\usepackage{stfloats}
\usepackage{url}
\usepackage{verbatim}
\usepackage{graphicx}
\usepackage{pdfpages}
\usepackage{cite}
\usepackage{xcolor}
\usepackage[numbers,sort&compress]{natbib}
\usepackage{booktabs}
\usepackage{multirow}
\usepackage{lipsum}
\usepackage{hyperref}
\usepackage[edges]{forest}
\definecolor{hidden-draw}{RGB}{20,68,106}
\definecolor{hidden-pink}{RGB}{255,245,247}
\usepackage[framemethod=tikz]{mdframed}
\usepackage{subcaption}
\usepackage{soul}
\usepackage{times}
\usepackage{tabularx}
\usepackage{float}
\usepackage{xcolor} 
\usepackage{latexsym}
\usepackage{bbm}
\usepackage{amsfonts}
\usepackage{dsfont}
\usepackage{graphicx}
\usepackage{caption}
\usepackage{subcaption}
\usepackage{algpseudocode}
\usepackage{colortbl}
\definecolor{lightgray}{gray}{0.9}
\definecolor{lightgreen}{rgb}{0.9, 1, 0.9}
\usepackage{microtype}
\usepackage{inconsolata}
\usepackage[edges]{forest}

\newcommand{\TA}{TA}

\begin{document}

\title{Next-Generation Database Interfaces: \\A Survey of LLM-based Text-to-SQL}

\author{Zijin~Hong,
        Zheng~Yuan,
        Qinggang~Zhang\IEEEauthorrefmark{2},
        Hao~Chen,
        Junnan~Dong, \\
        Feiran~Huang,~\IEEEmembership{Senior Member,~IEEE},
        and Xiao~Huang
\thanks{This work was supported in part by the Research Grants Council of the Hong Kong Special Administrative Region, China under Project PolyU under Grant 25208322, in part by NSFC under Grant 62502008, and in part by the Funding Scheme for Research and Innovation of FDCT under Grant 0021/2025/ITP1.
(\IEEEauthorrefmark{2}\textit{Corresponding author: Qinggang~Zhang.)}}
\IEEEcompsocitemizethanks{
    \IEEEcompsocthanksitem Zijin~Hong, Zheng~Yuan, Qinggang~Zhang, Junnan~Dong and Xiao~Huang are with the Department of Computing, The Hong Kong Polytechnic University, Hong Kong SAR, China. E-mail: \{zijin.hong, yzheng.yuan, hanson.dong\}@connect.polyu.hk; \{qinggang.zhang, xiao.huang\}@polyu.edu.hk
    \IEEEcompsocthanksitem Hao~Chen is with the Faculty of Data Science, City University of Macau, Macao SAR, China. E-mail: sundaychenhao@gmail.com.
    \IEEEcompsocthanksitem Feiran~Huang is with the College of Cyber Security, Jinan University, Guangzhou, China. E-mail: huangfr@jnu.edu.cn.
    }
}

\maketitle

\begin{abstract}
Generating accurate SQL from users' natural language questions (text-to-SQL) remains a long-standing challenge due to the complexities involved in user question understanding, database schema comprehension, and SQL generation. 
Traditional text-to-SQL systems, which combine human engineering and deep neural networks, have made significant progress. 
Subsequently, pre-trained language models (PLMs) have been developed for text-to-SQL tasks, achieving promising results. 
However, as modern databases and user questions grow more complex, PLMs with a limited parameter size often produce incorrect SQL. 
This necessitates more sophisticated and tailored optimization methods, which restrict the application of PLM-based systems. 
Recently, large language models (LLMs) have shown significant capabilities in natural language understanding as model scale increases. 
Thus, integrating LLM-based solutions can bring unique opportunities, improvements, and solutions to text-to-SQL research. 
In this survey, we provide a comprehensive review of existing LLM-based text-to-SQL studies. 
Specifically, we offer a brief overview of the technical challenges and evolutionary process of text-to-SQL. 
Next, we introduce the datasets and metrics designed to evaluate text-to-SQL systems. 
Subsequently, we present a systematic analysis of recent advances in LLM-based text-to-SQL. 
Finally, we make a summary and discuss the remaining challenges in this field and suggest expectations for future research directions. 
All the related resources of LLM-based, including research papers, benchmarks, and open-source projects, are collected for the community in our repository: \textcolor{blue}{\url{https://github.com/DEEP-PolyU/Awesome-LLM-based-Text2SQL}}.

\end{abstract}


\ifCLASSOPTIONcompsoc
\IEEEraisesectionheading{\section{Introduction}\label{sec:introduction}}
\else
\section{Introduction}
\label{sec:introduction}
\fi
\begin{figure}[!t]
    \centering
    \includegraphics[width=\linewidth]{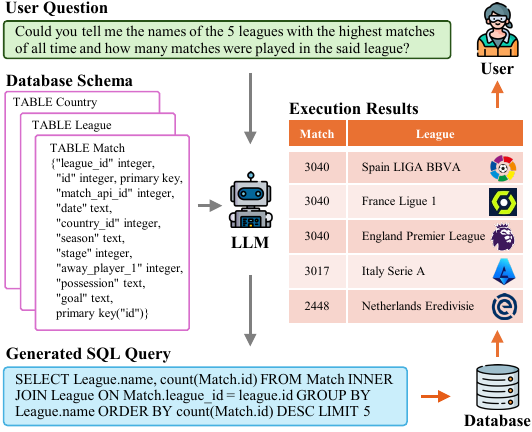}
    \caption{An example of LLM-based text-to-SQL is selected from the BIRD~\cite{li2023BIRD} dataset. A user asks a question about football leagues. The LLM takes this question along with the schema of the corresponding database as input and generates an SQL query as output. The generated SQL query can be executed in the database, retrieves the content \textit{``The 5 leagues with the highest matches''}, providing the answer to the user's question.
    }
    \label{fig:example}
\end{figure}
\IEEEPARstart{T}{ext-to-SQL} is a long-standing task in natural language processing, aiming to convert natural language (NL) questions into database-executable SQL queries. 
Fig.~\ref{fig:example} illustrates an example of a large language model-based (LLM-based) text-to-SQL system. 
Given a user question such as \textit{``Could you tell me the names of the 5 leagues with the highest matches of all time and how many matches were played in the said league?''}, the LLM takes the question and its corresponding database schema as input to generate an SQL query. 
Executing this SQL query in the database retrieves the relevant content needed to answer the user's question. 
This system serves as a natural language interface to the database (NLIDB) using LLMs. 
According to the 2023 Stack Overflow survey, SQL is one of the most widely used programming languages, with over half of professional developers (51.52\%) employing it, while only about a third (35.29\%) having received systematic training.
The NLIDB therefore allows non-skilled users to access structured databases like professional database engineers~\cite{wang2022proton,qin2022survey} and enhances human-computer interaction~\cite{xu2020autoqa}. 
Furthermore, amid the research hotspot of LLMs, text-to-SQL can potentially mitigate prevalent hallucination issues by integrating realistic content from databases to fill LLMs' knowledge gaps~\cite{lin2021truthfulqa}. 
The significant value of text-to-SQL has triggered a range of studies on its integration and optimization with LLMs~\cite{rajkumar2022evaluating,pourreza2023dinsql,gao2023dailsql}; consequently, LLM-based text-to-SQL remains a highly discussed research field in the NLP and database communities.

\begin{figure*}[!t]
    \centering
    \includegraphics[width=\linewidth]{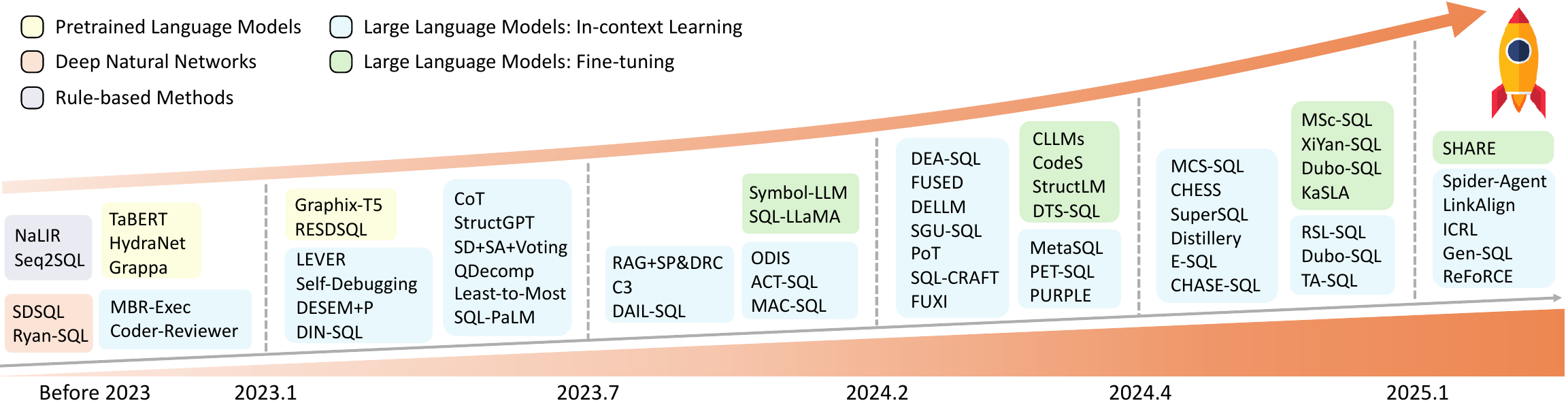}
    \caption{A sketch of text-to-SQL research evolution outlines the advancement of implementation paradigms, each represented with different color backgrounds. Before 2023, the focus is on a selection of representative traditional studies. However, from 2023 onward, the emphasis shifts to the rapid advancements driven by LLMs, marking a significant acceleration in the field.}
    \label{fig:milestone}
\end{figure*}

Previous traditional methods have made notable progress in implementing text-to-SQL. 
As shown in Fig.~\ref{fig:milestone}, these implementations have undergone a long evolutionary process. 
Early efforts were mostly based on well-designed rules and templates~\cite{li2014constructing}, specifically suitable for simple database scenarios. 
In recent years, due to the heavy labor costs~\cite{mahmud2015rule} associated with rule-based methods and the growing complexity of database environments~\cite{yu2018Spider,zhong2017WikiSQL,pourreza2023evaluating}, designing a rule or template for each scenario has become increasingly difficult and impractical. 
The development of deep neural networks has advanced the progress of text-to-SQL~\cite{sutskever2014sequence,vaswani2017attention}, which can automatically learn a mapping from the user question to its corresponding SQL~\cite{hui2021improving,choi2021ryansql}. 
Subsequently, pre-trained language models (PLMs) with strong semantic parsing capabilities have become the new paradigm for text-to-SQL systems~\cite{yin2020tabert}, boosting their performance to a new level~\cite{li2023resdsql,li2023graphix,rai2023improving}. 
Incremental research on PLM-based optimization, such as table content encoding~\cite{yin2020tabert,choi2021ryansql,lyu2020hybrid} and pre-training~\cite{yin2020tabert,yu2021grappa}, has further advanced this field. 
Most recently, LLM-based approaches have implemented text-to-SQL through the in-context learning~\cite{pourreza2023dinsql} and fine-tuning~\cite{li2024codes} paradigms, reaching state-of-the-art accuracy with well-designed frameworks and stronger comprehension capabilities compared to PLM-based methods.

The overall implementation details of LLM-based text-to-SQL can be divided into three aspects:  
\textbf{1. Question understanding}: The NL question is a semantic representation of the user's intention, which the corresponding generated SQL query is expected to align with;
\textbf{2. Schema comprehension}: The schema provides the table and column structure of the database, and the text-to-SQL system is required to identify the target components that match the user question;
\textbf{3. SQL generation}: This involves incorporating the above parsing and then predicting the correct syntax to generate executable SQL queries that can retrieve the required answer. 
LLMs have proven to perform a successful vanilla implementation~\cite{rajkumar2022evaluating,liu2023comprehensive}, benefiting from the powerful semantic parsing capacity enabled by a rich training corpus~\cite{yang2024harnessing,zhao2023survey}. Further studies on enhancing LLMs for question understanding~\cite{pourreza2023dinsql,gao2023dailsql}, schema comprehension~\cite{dong2023c3,yuan2025knapsack}, and SQL generation~\cite{zhang2024sgusql} are being increasingly released.

\tikzstyle{my-box}=[
    rectangle,
    draw=hidden-draw,
    rounded corners,
    align=left,
    text opacity=1,
    minimum height=1.5em,
    minimum width=5em,
    inner sep=2pt,
    fill opacity=.8,
    line width=0.8pt,
]
\tikzstyle{leaf-head}=[my-box, minimum height=1.5em,
    draw=gray!80,
    fill=gray!15,
    text=black, font=\normalsize,
    inner xsep=2pt,
    inner ysep=4pt,
    line width=0.8pt,
]

\tikzstyle{leaf-datasets}=[my-box, minimum height=1.5em,
    draw=red!70,
    fill=red!15,
    text=black, font=\normalsize,
    inner xsep=2pt,
    inner ysep=4pt,
    line width=0.8pt,
]

\tikzstyle{leaf-methods}=[my-box, minimum height=1.5em,
    draw=cyan!70,
    fill=cyan!15,
    text=black, font=\normalsize,
    inner xsep=2pt,
    inner ysep=4pt,
    line width=0.8pt,
]
\tikzstyle{leaf-metrics}=[my-box, minimum height=1.5em,
    draw=orange!80,
    fill=orange!15,
    text=black, font=\normalsize,
    inner xsep=2pt,
    inner ysep=4pt,
    line width=0.8pt,
]

\tikzstyle{modelnode-datasets}=[my-box, minimum height=1.5em,
    draw=red!80,
    fill=white,
    text=black, font=\normalsize,
    inner xsep=2pt,
    inner ysep=4pt,
    line width=0.8pt,
]

\tikzstyle{modelnode-methods}=[my-box, minimum height=1.5em,
    draw=cyan!100,
    fill=white,
    text=black, font=\normalsize,
    inner xsep=2pt,
    inner ysep=4pt,
    line width=0.8pt,
]
\tikzstyle{modelnode-metrics}=[my-box, minimum height=1.5em,
    draw=orange!100,
    fill=white,
    text=black, font=\normalsize,
    inner xsep=2pt,
    inner ysep=4pt,
    line width=0.8pt,
]

\begin{figure*}[!th]
    \centering
    \resizebox{1\textwidth}{!}{
        \begin{forest}
            forked edges,
            for tree={
                grow=east,
                reversed=true,
                anchor=base west,
                parent anchor=east,
                child anchor=west,
                base=left,
                font=\normalsize,
                rectangle,
                draw=hidden-draw,
                rounded corners,
                align=left,
                minimum width=1em,
                edge+={darkgray, line width=1pt},
                s sep=3pt,
                inner xsep=0pt,
                inner ysep=3pt,
                line width=0.8pt,
                ver/.style={rotate=90, child anchor=north, parent anchor=south, anchor=center},
            }, 
            [
                LLM-based Text-to-SQL, leaf-head, ver
                [
                    Datasets \\ (\S\ref{sec:dataset}), leaf-datasets, text width=5em
                    [
                        Original \\ Datasets \\ \& \\ Post-annotated \\ Datasets, leaf-datasets, text width=6em
                        [
                            Cross- \\ domain, leaf-datasets, text width=5.5em
                            [BIRD~\cite{li2023BIRD}{, }DuSQL~\cite{wang2020DuSQL}{, }CoSQL~\cite{yu2019CoSQL}{, }Spider~\cite{yu2018Spider}{, }WikiSQL~\cite{zhong2017WikiSQL}{, }KaggleDBQA~\cite{lee2021KaggleDBQA}{, }ADVETA~\cite{pi2022ADVETA}{, } \\ Spider-SS~\cite{gan2022Spider-CG}{, }Spider-CG~\cite{gan2022Spider-CG}{, }Spider-DK~\cite{gan2021Spider-DK}{, }Spider-SYN~\cite{gan2021Spider-SYN}{, }Spider-Realistic~\cite{deng2021Spider-Realistic}{, }CSpider~\cite{min2019CSpider}{, } \\ SParC~\cite{yu2019SParC}, modelnode-datasets, text width=43.5em]
                        ]
                        [
                            Knowledge- \\ augmented, leaf-datasets, text width=5.5em
                            [BIRD~\cite{li2023BIRD}{, }SQUALL~\cite{shi2020SQUALL}{, }Spider-DK~\cite{gan2021Spider-DK}{, }BIRD-CRITIC~\cite{li2023BIRD}{, }Spider 2.0~\cite{lei2025spider2}, modelnode-datasets, text width=43.5em]
                        ]
                        [
                            Cross- \\ lingual, leaf-datasets, text width=5.5em
                            [DuSQL~\cite{wang2020DuSQL}{, }CSpider~\cite{min2019CSpider}{, }Spider-Vietnamese~\cite{tuan2020pilot}, modelnode-datasets, text width=43.5em]
                        ]
                        [
                            Context- \\ dependent, leaf-datasets, text width=5.5em
                            [CoSQL~\cite{yu2019CoSQL}{, }Spider-SS~\cite{gan2022Spider-CG}{, }Spider-CG~\cite{gan2022Spider-CG}{, }SparC~\cite{yu2019SParC}, modelnode-datasets, text width=43.5em]
                        ]
                        [
                            Robustness, leaf-datasets, text width=5.5em
                            [ADVETA~\cite{pi2022ADVETA}{, }Spider-SYN~\cite{gan2021Spider-SYN}{, }Spider-Realistic~\cite{deng2021Spider-Realistic}{, }Dr. Spider~\cite{chang2023dr}, modelnode-datasets, text width=43.5em]
                        ]
                        [
                            Long-context, leaf-datasets, text width=5.5em
                            [BIRD-CRITIC~\cite{li2023BIRD}{, }Spider 2.0~\cite{lei2025spider2}, modelnode-datasets, text width=43.5em]
                        ]
                        [
                            Specialized- \\ domain, leaf-datasets, text width=5.5em
                            [BULL~\cite{zhang2024finsql}, modelnode-datasets, text width=43.5em]
                        ]
                    ]
                ]
                [
                    Evaluation \\ Metrics \\ (\S\ref{sec:metric}), leaf-metrics,text width=5em
                    [ 
                        Content- \\ Matching \\based, leaf-metrics, text width=6em
                        [Component Matching (CM)~\cite{yu2018Spider}{, }Exact Matching (EM)~\cite{yu2018Spider}, modelnode-metrics, text width=50.7em]
                    ]
                    [ 
                        Execution \\based, leaf-metrics, text width=6em
                        [Execution Accuracy (EX)~\cite{yu2018Spider}{, }Valid Efficiency Score (VES)~\cite{li2023BIRD}, modelnode-metrics, text width=50.7em]
                    ]
                ]
                [
                    Methods \\ (\S\ref{sec:method}), leaf-methods,text width=5em
                    [
                        In-context \\ Learning \\ Paradigm \\ (\S\ref{sec:icl}), leaf-methods, text width=6em
                        [
                            Decomposition, leaf-methods, text width=6.5em
                            [Coder-Reviewer~\cite{zhang2022coder}{, }DIN-SQL~\cite{pourreza2023dinsql}{, }QDecomp~\cite{tai2023exploring}{, }C3~\cite{dong2023c3}{, }MAC-SQL~\cite{wang2024macsql}{, }DEA-SQL~\cite{xie2024deasql}{, } \\ SGU-SQL~\cite{zhang2024sgusql}{, }MetaSQL~\cite{fan2024metasql}{, }$\textit{\textbf{R}}^3$~\cite{xia2024r3}{, }PET-SQL~\cite{li2024petsql}{, }PURPLE~\cite{ren2024purple}{, }TA-SQL~\cite{qu2024generation}{, }MCS-SQL~\cite{lee2025mcs}{, } \\ CHESS~\cite{talaei2024chess}{, }SuperSQL~\cite{li2024dawn}{, }Distillery~\cite{maamari2024the}{, }RSL-SQL~\cite{cao2024rsl}{, }Gen-SQL~\cite{shi2025gen}{, }Spider-Agent~\cite{lei2025spider2}{, } \\ ReFoRCE~\cite{deng2025reforce}{, }LinkAlign~\cite{wang2025linkalign}, modelnode-methods,text width=42.5em]
                        ]
                        [
                            Prompt \\ Optimization, leaf-methods, text width=6.5em
                            [DESEM+P~\cite{guo2023desem}{, }StructGPT~\cite{jiang2023structgpt}{, }SD+SA+Voting~\cite{nan2023enhancing}{, }RAG+SP\&DRC~\cite{guo2023ragsql}{, }C3~\cite{dong2023c3}{, }DAIL-SQL~\cite{gao2023dailsql}{, } \\ ODIS~\cite{chang2023selective}{, }ACT-SQL~\cite{zhang2023actsql}{, }FUSED~\cite{wang2024fused}{, }DELLM~\cite{hong2024knowledgetosql}{, }Dubo-SQL~\cite{thorpe2024dubo}{, }\TA-SQL~\cite{qu2024generation}{, } \\ MCS-SQL~\cite{lee2025mcs}{, }  SuperSQL~\cite{li2024dawn}{, }Distillery~\cite{maamari2024the}{, }RSL-SQL~\cite{cao2024rsl}{, }ICRL~\cite{toteja2025context}{, }Gen-SQL~\cite{shi2025gen}{, } \\ SAFE-SQL~\cite{lee2025safe}{, }  ReFoRCE~\cite{deng2025reforce}{, }LinkAlign~\cite{wang2025linkalign}, modelnode-methods,text width=42.5em]
                        ]
                        [
                            Reasoning \\ Enhancement, leaf-methods, text width=6.5em[CoT~\cite{tai2023exploring,li2023BIRD,gao2023dailsql,zhang2024sgusql}{, }QDecomp~\cite{tai2023exploring}{, }Least-to-Most~\cite{tai2023exploring}{, }SQL-PaLM~\cite{sun2023sqlpalm}{, }ACT-SQL~\cite{zhang2023actsql}{, }POT~\cite{xia2024sqlcraft}{, } \\ SQL-CRAFT~\cite{xia2024sqlcraft}{, }FUXI~\cite{gu2024fuxi}{, }SuperSQL~\cite{li2024dawn}{, }CHASE-SQL~\cite{pourreza2025chasesql}{, }ReFoRCE~\cite{deng2025reforce}{, }Spider-Agent~\cite{lei2025spider2}, modelnode-methods,text width=42.5em]
                        ]
                        [
                            Execution \\ Refinement, leaf-methods, text width=6.5em[MBR-Exec~\cite{shi2022natural}{, }Coder-Reviewer~\cite{zhang2022coder}{, }LEVER~\cite{ni2023lever}{, }SELF-DEBUGGING~\cite{chen2024selfdebugging}{, }DESEM+P~\cite{guo2023desem}{, } \\ DIN-SQL~\cite{pourreza2023dinsql}{, }SD+SA+Voting~\cite{nan2023enhancing}{, }SQL-PaLM~\cite{sun2023sqlpalm}{, }RAG+SP\&DRC~\cite{guo2023ragsql}{, }C3~\cite{dong2023c3}{, }MAC-SQL~\cite{wang2024macsql}{, } \\ DELLM~\cite{hong2024knowledgetosql}{, }SQL-CRAFT~\cite{xia2024sqlcraft}{, }FUXI~\cite{gu2024fuxi}{, }$\textit{\textbf{R}}^3$~\cite{xia2024r3}{, }PET-SQL~\cite{li2024petsql}{, }PURPLE~\cite{ren2024purple}{, }Dubo-SQL~\cite{thorpe2024dubo}{, } \\ MCS-SQL~\cite{lee2025mcs}{, }CHESS~\cite{talaei2024chess}{, }Distillery~\cite{maamari2024the}{, }E-SQL~\cite{caferouglu2024esql}{, }CHASE-SQL~\cite{pourreza2025chasesql}{, }RSL-SQL~\cite{cao2024rsl}{, } \\ Spider-Agent~\cite{lei2025spider2}{, }ReFoRCE~\cite{deng2025reforce}, modelnode-methods,text width=42.5em]
                        ]
                    ]
                    [
                        Fine-tuning \\ Paradigm \\ (\S\ref{sec:ft}), leaf-methods,text width=6em
                        [
                            Enhanced \\ Architecture, leaf-methods, text width=6.5em
                            [CLLMs~\cite{kou2024cllms}, modelnode-methods,text width=42.5em]
                        ]
                        [
                            Pre-training, leaf-methods, text width=6.5em
                            [CodeS~\cite{li2024codes}, modelnode-methods,text width=42.5em]
                        ]
                        [
                            Data \\ Augmentation, leaf-methods, text width=6.5em
                            [DAIL-SQL~\cite{gao2023dailsql}{, }Symbol-LLM~\cite{xu2024symbolllm}{, }CodeS~\cite{li2024codes}{, }StructLM~\cite{zhuang2024structlm}{, }Dubo-SQL~\cite{thorpe2024dubo}{, }Distillery~\cite{maamari2024the}{, } \\ XiYan-SQL~\cite{gao2024xiyan}{, }SHARE~\cite{qu2025share}, modelnode-methods,text width=42.5em]
                        ]
                        [
                            Multi-task \\ Tuning, leaf-methods, text width=6.5em
                            [SQL-LLaMA~\cite{roziere2023codellama}{, }DTS-SQL~\cite{pourreza2024dtssql}{, }KaSLA~\cite{yuan2025knapsack}{, }MSc-SQL~\cite{gorti2024mscsql}{, }XiYan-SQL~\cite{gao2024xiyan}{, }ROUTE~\cite{qin2025route}{, } \\ SHARE~\cite{qu2025share}, modelnode-methods,text width=42.5em]
                        ]
                    ]
                ]  
            ]
        \end{forest}
    }
    \caption{Taxonomy tree of research in LLM-based text-to-SQL, taking inspiration from \cite{xu2023large}; nodes are ordered by release time.}
    \label{fig:taxonomy}
\end{figure*}

Despite the significant progress made in text-to-SQL research, several challenges remain that hinder the development of robust and generalized text-to-SQL systems~\cite{katsogiannis2023survey}. 
Recent related works have surveyed text-to-SQL systems using deep learning approaches and provided insights into previous deep neural network and PLM-based research~\cite{deng2022recent,qin2022survey,zhao2023survey,zhang2024tabluarsurvey}. 
In this survey, we aim to catch up with recent advances and provide a comprehensive review of the current state-of-the-art models and approaches in LLM-based text-to-SQL. 
We begin by introducing the fundamental concepts and challenges associated with text-to-SQL, highlighting the importance of this task in various domains. 
We then delve into the evolution of the implementation paradigm for text-to-SQL systems, discussing the key advancements and breakthroughs in this field. 
After the overview, we provide a detailed introduction and analysis of the recent advances in text-to-SQL integrating LLMs. 
Specifically, the body of our survey covers a range of topics related to LLM-based text-to-SQL, including:
\begin{itemize}
    \item \textbf{Benchmarks and Metrics:} We introduce the commonly used datasets and benchmarks for evaluating LLM-based text-to-SQL systems. We discuss their characteristics, complexity, and the challenges they pose for development and evaluation, along with the evaluation metrics.
    \item \textbf{Methods:} We systematically analyze different methods proposed for LLM-based text-to-SQL, including in-context learning and fine-tuning paradigms. We explore their implementation details, strengths, and specific adaptations, organized through well-designed categorization.
    \item \textbf{Summarization and Expectations:} We summarize the advantages and limitations of both LLM-based and traditional methods, as well as within LLM-based methods themselves. We discuss remaining challenges such as real-world robustness, computational efficiency, data privacy, and possible extensions. Additionally, we outline future research directions and opportunities for improvement.
\end{itemize}

To summarize our technical contributions, \textbf{at the time of this paper's release, we present the first LLM-based text-to-SQL survey in the community}, establishing a systematic taxonomy for LLM-based text-to-SQL methods. 
\textbf{This categorization provides analysis from technical implementation perspectives}, clearly indicating their contributions to generating better SQL. 
Then, \textbf{we further integrate insights while analyzing the trade-offs} between LLM-based and traditional methods, as well as between in-context learning and fine-tuning paradigms for future research. 
Finally, \textbf{we highlight the remaining challenges} in robustness, real-world deployment, efficiency, and domain adaptation, \textbf{offering actionable directions for future research}.
Fig.~\ref{fig:taxonomy} shows a taxonomy for the survey.

\section{Overview}
Text-to-SQL is a task that aims to convert NL questions into SQL queries that can be executed in a relational database. 
It has the potential to democratize data access by allowing users to interact with databases using natural language, thus eliminating the need for specialized SQL programming knowledge~\cite{ma2021mt}. 
This capability can greatly benefit domains such as business intelligence, customer support, and scientific research by enabling non-specialized users to easily retrieve target content from databases and facilitating more efficient data analysis.

\subsection{Challenges in Text-to-SQL}
The technical challenges are summarized as follows: 

\subsubsection{Linguistic Complexity and Ambiguity} 
Natural language questions often contain complex linguistic representations, such as nested clauses, coreferences, and ellipses, which make it challenging to map them accurately to the corresponding parts of SQL queries~\cite{deng2021Spider-Realistic}. 
Additionally, natural language is inherently ambiguous, with multiple possible representations for a given user question~\cite{rajpurkar2016squad,rajpurkar2018know}. 
Resolving these ambiguities and understanding the intent behind the user question requires deep natural language understanding and the capability to incorporate context and domain knowledge~\cite{li2023BIRD}.


\subsubsection{Schema Understanding and Representation}
To generate accurate SQL queries, text-to-SQL systems must have a comprehensive understanding of the database schema, including table names, column names, and relationships between various tables. 
Representing and encoding schema information is challenging because the database schema is often complex and varies significantly across different domains~\cite{yu2018Spider}.


\subsubsection{Rare and Complex SQL Operations}
Some SQL queries involve rare or complex operations and syntax in challenging scenarios, such as nested sub-queries, outer joins, and window functions. 
These operations are less frequent in the training data and pose challenges for text-to-SQL systems to generate accurately. 
Designing models that can generalize to a wide range of SQL operations, including rare and complex scenarios, is an essential consideration.

\subsubsection{Cross-Domain Generalization}
Text-to-SQL systems often struggle to generalize across various database scenarios and domains. 
Models trained on a specific domain may not perform well on questions from other domains due to variations in vocabulary, database schema structure, and question patterns. 
Developing systems that can effectively generalize to new domains with minimal domain-specific training data or fine-tuning adaptation is a significant challenge~\cite{yang2024unveiling}.

\subsection{Evolutionary Process}
The research field of text-to-SQL has witnessed significant advancements over the years in the NLP community, having evolved from rule-based methods to deep learning-based approaches and, more recently, to integrating pre-trained language models (PLMs) and large language models (LLMs), a sketch of the evolutionary process is shown in Fig.~\ref{fig:milestone}.

\subsubsection{Rule-based Methods}
Early text-to-SQL systems relied heavily on rule-based methods~\cite{li2014constructing,mahmud2015rule,yu2021grappa}, where manually crafted rules and heuristics were used to map natural language questions to SQL queries. 
Rule-based systems excel at generating syntactically correct SQL queries by relying on predefined templates and strict grammatical rules, ensuring adherence to SQL syntax and structure. Their deterministic nature guarantees consistent outputs for specific input patterns, minimizing variability in query generation.
However, these methods struggle to interpret linguistically complex or ambiguous natural language questions, such as those involving nested clauses, coreferences, or ellipses, and often fail to map these elements to database schema components like tables or columns. Their reliance on manual rules limits generalization to unseen question patterns or schema structures, requiring frequent updates for new domains. Additionally, encoding complex schemas with multi-table relationships into usable rules is labor-intensive and prone to errors.

\subsubsection{Deep Learning-based Approaches}
With the rise of deep neural networks, sequence-to-sequence models and encoder-decoder structures, such as LSTMs~\cite{hochreiter1997long} and transformers~\cite{vaswani2017attention}, were adapted to generate SQL queries from natural language input~\cite{guo2019towards,choi2021ryansql}. 
Typically, RYANSQL~\cite{choi2021ryansql} introduced techniques like intermediate representations and sketch-based slot filling to handle complex questions and improve cross-domain generalization. 
Recently, researchers introduced graph neural networks (GNNs) for text-to-SQL tasks by leveraging schema dependency graphs to capture the relationships between database elements~\cite{hui2021improving,xu2017sqlnet}.
Despite their flexibility, these methods often generate incorrect or incomplete SQL syntax, such as missing clauses or improper nesting, due to limited structural awareness. They also struggle with rare or complex SQL operations, like nested subqueries and window functions, because such constructs are underrepresented in training data.

\subsubsection{PLM-based Implementation}
PLMs have emerged as a powerful solution for text-to-SQL, leveraging the vast amounts of linguistic knowledge and semantic understanding captured during the pre-training process. 
The early adoption of PLMs in text-to-SQL primarily focused on fine-tuning off-the-shelf PLMs, such as BERT~\cite{devlin2019bert} and RoBERTa~\cite{liu2019roberta}, on standard text-to-SQL datasets~\cite{yu2018Spider,zhong2017WikiSQL}. 
These PLMs, pre-trained on large amounts of training corpus, captured rich semantic representations and language understanding capabilities. 
By fine-tuning them on text-to-SQL tasks, researchers aimed to leverage the semantic and linguistic understanding of PLMs to generate accurate SQL queries~\cite{guo2019towards,yin2020tabert,dou2022towards}.
Another line of research focuses on incorporating schema information into PLMs to improve their understanding of database structures and enable them to generate more executable SQL queries.
Schema-aware PLMs are designed to capture the relationships and constraints present in the database structure~\cite{li2023resdsql}.
While PLMs generate more accurate SQL than deep learning-based methods, they still face difficulties with complex operations like outer joins or aggregations, often producing syntactically flawed queries. Besides that, cross-domain performance drops significantly when schema structures or vocabulary differ from training data, necessitating resource-intensive fine-tuning.

\subsubsection{LLM-based Implementation}
Large language models (LLMs), such as the GPT series~\cite{radford2018improving,brown2020gpt3,achiam2023gpt4}, have gained significant attention in recent years due to their ability to generate coherent and fluent text. 
Researchers have begun exploring the potential of LLMs for text-to-SQL by leveraging their extensive knowledge reserves and superior generation capabilities~\cite{rajkumar2022evaluating,gao2023dailsql}. 
These approaches often involve prompt engineering to guide proprietary LLMs in SQL generation~\cite{chang2023how} or fine-tuning open-source LLMs on text-to-SQL datasets~\cite{gao2023dailsql}.

The integration of LLMs in text-to-SQL is still an emerging research area with significant potential for further exploration and improvement. 
Researchers are investigating ways to better leverage the knowledge and reasoning capabilities of LLMs, incorporate domain-specific knowledge~\cite{li2023BIRD,hong2024knowledgetosql}, and develop more efficient fine-tuning strategies~\cite{li2024codes}. 
As the field continues to evolve, we anticipate the development of more advanced and superior LLM-based implementations that will elevate the performance and generalization of text-to-SQL to new heights.

\section{Benchmarks \& evalution}
In this section, we introduce the benchmarks for text-to-SQL, encompassing well-known datasets and evaluation metrics. 

\begin{table*}[!t]
\caption{The statistics and analysis of well-known datasets of text-to-SQL ordered by release time. The original dataset indicates that the dataset is designed with a corresponding database, while post-annotated datasets involve annotating new components within existing datasets and databases rather than releasing a new database. For each dataset, the languages utilized are denoted by the following abbreviations: EN for English, ZH for Chinese, and VI for Vietnamese.}
\label{tab:datasets}
\centering
\resizebox{\linewidth}{!}{
\begin{tabular}{c|c|cccc|c|c}
\toprule
Original Dataset & Release Time & \#Example &\#DB &\#Table/DB &\#Row/DB &Language&Characteristics\\
\midrule
BIRD-CRITIC~\cite{li2023BIRD} &Feb-2025 &600 &- &- &- &EN&Knowledge-augmented, Long-context \\
Spider 2.0~\cite{lei2025spider2} &Nov-2024 &632 &213 &- &- &EN&Knowledge-augmented, Long-context \\
BULL~\cite{zhang2024finsql} &Jan-2024 &4,966&3&26&-&EN&Specialized-domain\\
BIRD~\cite{li2023BIRD} &May-2023 &12,751&95&7.3&549K&EN&Cross-domain, Knowledge-augmented\\
KaggleDBQA~\cite{lee2021KaggleDBQA} &Jun-2021 &272&8&2.3&280K&EN&Cross-domain \\
DuSQL~\cite{wang2020DuSQL} &Nov-2020 &23,797&200&4.1 &-&EN&Cross-domain, Cross-lingual\\ 
SQUALL~\cite{shi2020SQUALL} &Oct-2020 &11,468&1,679&1&-&EN&Knowledge-augmented \\ 
CoSQL~\cite{yu2019CoSQL} &Sep-2019 &15,598&200&-&-&EN&Cross-domain, Context-dependent \\
Spider~\cite{yu2018Spider} &Sep-2018&10,181&200&5.1&2K&EN&Cross-domain \\
WikiSQL~\cite{zhong2017WikiSQL} &Aug-2017 &80,654&26,521&1&17&EN&Cross-domain\\
\midrule
 Post-annotated Dataset& Release Time & \multicolumn{1}{c|}{Source Dataset} & \multicolumn{3}{c|}{Special Setting}&Language &Characteristics\\
\midrule
Dr. Spider~\cite{chang2023dr} &Jan-2023 &\multicolumn{1}{c|}{Spider}&\multicolumn{3}{c|}{Perturbations in DB, query and SQL}&EN&Robustness \\
ADVETA~\cite{pi2022ADVETA} &Dec-2022 & \multicolumn{1}{c|}{Spider, etc.}&\multicolumn{3}{c|}{Adversarial table perturbation}&EN&Robustness \\
Spider-SS\&CG~\cite{gan2022Spider-CG} &May-2022 &\multicolumn{1}{c|}{Spider}&\multicolumn{3}{c|}{Splitting example into sub-examples}&EN&Context-dependent \\ 
Spider-DK~\cite{gan2021Spider-DK} &Sep-2021 &\multicolumn{1}{c|}{Spider}&\multicolumn{3}{c|}{Adding domain knowledge}&EN&Knowledge-augmented \\
Spider-SYN~\cite{gan2021Spider-SYN} &Jun-2021&\multicolumn{1}{c|}{Spider}&\multicolumn{3}{c|}{Manual synonym replacement}&EN&Robustness \\
Spider-Vietnamese~\cite{tuan2020pilot} &Nov-2020&\multicolumn{1}{c|}{Spider}&\multicolumn{3}{c|}{Vietnamese version of Spider}&VI&Cross-lingual \\
Spider-Realistic~\cite{deng2021Spider-Realistic} &Oct-2020 &\multicolumn{1}{c|}{Spider}&\multicolumn{3}{c|}{Removing column names in question}&EN&Robustness \\
CSpider~\cite{min2019CSpider} &Sep-2019 &\multicolumn{1}{c|}{Spider}&\multicolumn{3}{c|}{Chinese version of Spider}&ZH&Cross-lingual \\ 
SParC~\cite{yu2019SParC} &Jun-2019 &\multicolumn{1}{c|}{Spider}&\multicolumn{3}{c|}{Annotate conversational contents}&EN&Context-dependent \\
\bottomrule
\end{tabular}}
\end{table*}

\subsection{Datasets}\label{sec:dataset}
As shown in Tab.~\ref{tab:datasets}, we categorize the datasets into ``Original Datasets'' and ``Post-annotated Datasets'' based on whether they were released with the original dataset and databases or developed by adapting existing datasets and databases with special settings. 
For the original datasets, we offer a detailed analysis, including the number of examples, databases, tables per database, and rows per database. 
For post-annotated datasets, we identify their source dataset and describe the special settings applied to them. 
To highlight the potential opportunities of each dataset, we annotate them according to their characteristics. 
The annotations are listed in the rightmost column of Tab.~\ref{tab:datasets}, which we will discuss in detail below.

\subsubsection{Cross-domain Dataset} This refers to datasets where the background information of different databases comes from various domains. Since real-world text-to-SQL applications often involve databases from multiple domains, most original text-to-SQL datasets~\cite{zhong2017WikiSQL,yu2018Spider,yu2019CoSQL,wang2020DuSQL,lee2021KaggleDBQA,li2023BIRD} and post-annotated datasets~\cite{yu2019SParC,min2019CSpider,gan2021Spider-DK,gan2021Spider-SYN,deng2021Spider-Realistic,pi2022ADVETA,gan2022Spider-CG} are in the cross-domain setting to fit well with the requirements of cross-domain applications.

\subsubsection{Knowledge-augmented Dataset} Interest in incorporating domain-specific knowledge into text-to-SQL tasks has increased significantly in recent years. 
BIRD~\cite{li2023BIRD} employs human database experts to annotate each text-to-SQL sample with external knowledge, categorized into numeric reasoning knowledge, domain knowledge, synonym knowledge, and value illustration. 
Similarly, Spider-DK~\cite{gan2021Spider-DK} defines and adds five types of domain knowledge for a human-curated version of the Spider dataset~\cite{yu2018Spider}: SELECT columns mentioned by omission, simple inference required, synonym substitution in cell value words, one non-cell value word generates a condition, and easy to conflict with other domains. 
Both studies found that human-annotated knowledge significantly improves SQL generation performance for samples requiring external domain knowledge.
Additionally, SQUALL~\cite{shi2020SQUALL} manually annotates alignments between the words in NL questions and the entities in SQL, providing finer-grained supervision than other datasets.

\subsubsection{Context-dependent Dataset} SParC~\cite{yu2019SParC} and CoSQL~\cite{yu2019CoSQL} explore context-dependent SQL generation by constructing a conversational database querying system. 
Unlike traditional text-to-SQL datasets that only have a single question-SQL pair for one example, SParC decomposes the question-SQL examples in the Spider dataset into multiple sub-question-SQL pairs to construct a simulated and meaningful interaction, including interrelated sub-questions that aid SQL generation, and unrelated sub-questions that enhance data diversity. 
CoSQL, in comparison, involves conversational interactions in natural language, simulating real-world scenarios to increase complexity and diversity. 
Additionally, Spider-SS\&CG~\cite{gan2022Spider-CG} splits the NL question in the Spider dataset~\cite{yu2018Spider} into multiple sub-questions and sub-SQLs, demonstrating that training on these sub-examples can improve a text-to-SQL system's generalization ability on out-of-distribution samples.

\subsubsection{Robustness Dataset} Evaluating the accuracy of text-to-SQL systems with polluted or perturbed database contents (e.g., schema and tables) is crucial for assessing robustness. 
Spider-Realistic~\cite{deng2021Spider-Realistic} removes explicit schema-related words from the NL questions, while Spider-SYN~\cite{gan2021Spider-SYN} replaces them with manually selected synonyms. 
ADVETA~\cite{pi2022ADVETA} introduces adversarial table perturbation (ATP), which perturbs tables by replacing original column names with misleading alternatives and inserting new columns with high semantic associations but low semantic equivalency. 
Based on the generation ability of large-scale pre-trained language models, Dr. Spider~\cite{chang2023dr} provided 17 different perturbation strategies on databases, natural language queries and SQL statements to measure the robustness of text-to-SQL systems.
These perturbations lead to significant drops in accuracy, as a text-to-SQL system with low robustness may be misled by incorrect matches between tokens in NL questions and database entities.

\subsubsection{Cross-lingual Dataset} The SQL keywords, function names, table and column names are typically written in English, posing challenges for applications in other languages. 
CSpider~\cite{min2019CSpider} addresses this by translating the Spider dataset into Chinese, identifying new challenges in word segmentation and cross-lingual matching between Chinese questions and English database contents. 
DuSQL~\cite{wang2020DuSQL} introduces a practical text-to-SQL dataset with Chinese questions and database contents provided in both English and Chinese.
Spider-Vietnamese~\cite{tuan2020pilot} translates the Spider dataset into Vietnamese, 
empirically revealing several strategies to enhance text-to-SQL methods within the Vietnamese context.

\subsubsection{Long-context dataset}
In real-world database applications, numerous challenging scenarios demand complex SQL queries with significant token lengths, often exceeding 100 tokens and involving multiple intricate operations. Spider 2.0~\cite{lei2025spider2} exemplifies this complexity by presenting enterprise-level text-to-SQL problems that require advanced reasoning across various SQL queries and dialects. 
Distinct from earlier benchmarks, Spider 2.0 contains numerous real-world tasks sourced from diverse production databases such as BigQuery and Snowflake, featuring large-scale, industrial schemas with hundreds or even thousands of columns, nested structures, and multiple schemas per instance.
Each task requires navigating not only complex SQL logic, but also integrating information from project codebases, database metadata, and dialect-specific documentation, closely reflecting how data workflows operate in enterprise settings. Gold-standard queries in this benchmark average close to 150 tokens, far surpassing previous datasets in both length and complexity. Notably, all SQLs are carefully annotated and reviewed for both semantic fidelity and execution correctness, ensuring challenges are both realistic and robust. As a result, Spider 2.0 highlighted the substantial gap between present-day LLMs' capabilities and the demands of authentic, long-context database scenarios.
In parallel, BIRD-CRITIC\footnote{\url{https://bird-critic.github.io/}}~\cite{li2023BIRD} introduces a diagnostic dimension, comprising 600 tasks designed to assess the efficacy of LLMs in resolving real-world user issues across diverse SQL dialects. Together, these datasets push the boundaries of text-to-SQL research, emphasizing the challenges of complex, long-context SQL environments.

\subsubsection{Specialized-domain dataset}
Despite advancements in text-to-SQL research, there is still a shortage of specialized datasets tailored for domain-specific analysis. To address this gap, BULL~\cite{zhang2024finsql} introduces a dedicated text-to-SQL benchmark for the financial sector. The dataset encompasses databases related to funds, stocks, and macroeconomic data, offering a comprehensive platform to evaluate and enhance text-to-SQL methods in financial applications, effectively tackling the financial domain's unique text-to-SQL challenges.

\subsection{Evaluation Metrics}\label{sec:metric}
We introduce four widely used evaluation metrics for the text-to-SQL task as follows: Component Matching and Exact Matching, which are based on SQL content matching, and Execution Accuracy and Valid Efficiency Score, which are based on execution results.

\subsubsection{Content Matching-based Metrics}
SQL content matching metrics focus on comparing the predicted
SQL with the ground truth based on their structural and syntactic similarities.

\begin{itemize}
    \item \textbf{Component Matching (CM)}~\cite{yu2018Spider} evaluates the performance of text-to-SQL system by measuring the exact match between predicted and ground truth SQL components—SELECT, WHERE, GROUP BY, ORDER BY, and KEYWORDS—using the F1 score. 
    Each component is decomposed into sets of sub-components and compared for an exact match, accounting for SQL components without order constraints.
    \item \textbf{Exact Matching (EM)}~\cite{yu2018Spider} measures the percentage of examples whose predicted SQL query is identical to the ground truth SQL query. 
    A predicted SQL is considered correct only if all its components, as described in CM, match exactly with those of the ground truth query.
\end{itemize}

\subsubsection{Execution-based Metrics}
Execution result metrics assess the correctness of the generated SQL query by comparing the results obtained from executing the query on the target database with the expected results.
\begin{itemize}
    \item \textbf{Execution Accuracy (EX)}~\cite{yu2018Spider} measures the correctness of a predicted SQL query by executing it in the corresponding database and comparing the executed results with the results obtained by the ground truth query.
    \item \textbf{Valid Efficiency Score (VES)}~\cite{li2023BIRD} is defined to measure the efficiency of valid SQL queries. 
    A valid SQL query is a predicted SQL whose executed results exactly match the ground truth results. Specifically, VES evaluates both the efficiency and accuracy of predicted SQL queries. 
    For a text dataset with $N$ examples, VES is computed by:
        \begin{equation}
            \text{VES} = \frac{1}{N}\sum_{n=1}^{N}\mathbbm{1}(V_n, \hat{V}_n) \cdot \textbf{R}(Y_n, \hat{Y}_n),
        \end{equation}
    where $\hat{Y}_n$ and $\hat{V}_n$ are the predicted SQL query and its executed results and $Y_n$ and $V_n$ are the ground truth SQL query and its corresponding executed results, respectively.
    $\mathbbm{1}(V_n, \hat{V}_n)$ is an indicator function, where: 
    \begin{equation}
    \mathbbm{1}(V_n, \hat{V}_n) =
        \begin{cases}
         1, V_n = \hat{V}_n \\
         0, V_n \neq \hat{V}_n
        \end{cases}
    \end{equation}
    Then, $\textbf{R}(Y_n, \hat{Y}_n) = \sqrt{E(Y_n)/E(\hat{Y}_n)}$ denotes the relative execution efficiency of the predicted SQL in comparison to the ground-truth, where $E(\cdot)$ is the execution time of each SQL in the database. BIRD benchmark~\cite{li2023BIRD} ensures the stability of this metric by computing the average of $\textbf{R}(Y_n, \hat{Y}_n)$ over 100 runs for each example.
\end{itemize}
Most of the recent LLM-based text-to-SQL studies focus on BIRD and Spider-related datasets and execution-based metrics. The corresponding details can be found at Tab.~\ref{tab:icl-body}.

\section{Methods}\label{sec:method}
The implementation of the most current LLM-based text-to-SQL methods relies on in-context learning (ICL)~\cite{sahoo2024systematic,wang2023prompt,chen2023unleashing} and fine-tuning (FT)~\cite{wei2021finetuned,zheng2024llamafactory} paradigms due to the release of powerful proprietary and well-architected open-source LLMs in large quantities~\cite{achiam2023gpt4,xue2024dbgpt,wang2023dbcopilot,touvron2023llama,touvron2023llama2,bai2023qwen}.
In this section, we will discuss these paradigms with detailed categorization accordingly.

\subsection{In-context Learning}\label{sec:icl}
Through extensive and widely recognized research, ICL (also called prompt engineering) has been proven to play a decisive role in the performance of LLMs across various tasks~\cite{radford2019unsupervised,yang2024harnessing}.
Consequently, it also impacts SQL generation across different prompt styles~\cite{gao2023dailsql,zhang2024benchmarking}.
The ICL implementation of the LLM-based text-to-SQL process to generate a predicted SQL query $\hat{Y}$, can be formulated as:
\begin{equation}\label{eq:icl_process}
    \hat{Y} = \pi\left(I, Q, \mathcal{S} \mid \theta\right),
\end{equation}
where $I$ represents the instruction for the text-to-SQL task, 
$Q$ represents the user question. 
$\mathcal{S}$ is the database schema/content, which can be decomposed as $\mathcal{S} = \langle \mathcal{C},\mathcal{T}, \mathcal{K} \rangle$. 
Here, $\mathcal{C} = \{c_1, c_2, ...\}$ and $\mathcal{T} = \{t_1, t_2, ...\}$ represent the collection of various columns $c$ and tables $t$, respectively. 
$\mathcal{K}$ is the potentially supplementary knowledge, such as foreign key relationships~\cite{zhang2023actsql}, schema linking~\cite{dong2023c3,yuan2025knapsack}, and external knowledge~\cite{li2023BIRD,hong2024knowledgetosql}. 
$\pi(\cdot \mid \theta)$ is a LLM $\pi$ with parameter $\theta$. 
In the ICL paradigm, LLM-based methods primarily utilize an off-the-shelf text-to-SQL model, meaning the parameter $\theta$ of the model is frozen, for generating the predicted SQL query. 

We further specify the prompt engineering strategies in the ICL paradigm as zero-shot and few-shot prompting.
Trained through massive data, LLMs have a strong overall proficiency in different downstream tasks with zero-shot and few-shot prompting~\cite{reynolds2021prompt,wei2021finetuned,ye2022unreliability}, which is widely recognized and utilized in real-world applications.  
Formally, the overall input $P_{0}$ of zero-shot prompting for text-to-SQL can be obtained by concatenating $I$, $\mathcal{S}$, and $Q$ as introduced above:
\begin{equation}\label{eq:zero-shot}
    P_{0} = I \oplus \mathcal{S} \oplus Q.
\end{equation}
Subsequently, the overall input $P_{k}$ for few-shot prompting (specifically, $\text{$k$-shot}$) can be formulated as an extension of $P_{0}$:
\begin{equation}\label{eq:few-shot}
    P_{k} = \{\mathcal{E}_{1}, \mathcal{E}_{2},\ldots,\mathcal{E}_{k}\} \oplus P_{0},
\end{equation}
where $k$ is the number of provided instances (examples) in few-shot prompting. 
$\mathcal{E}_{i}$ represents the $i$-th few-shot instance, each of which can be decomposed as $\mathcal{E}_{i} = \left(\mathcal{S}_{i}, Q_{i}, Y_{i} \right)$, where $Y_i$ is the gold SQL query of $Q_{i}$. 
The index $i \in \{1,\ldots,k\}$ also serves as the corresponding serial number for $\mathcal{S}_i$, $Q_{i}$, and $Y_{i}$.
The OpenAI demonstration\footnote{The prompt format follows the official document provided by the OpenAI platform:~\url{https://platform.openai.com/examples/default-sql-translate}.} provides an indicative example and regulation for the prompting process in LLM-based text-to-SQL~\cite{dong2023c3}. 
The prompt designs and formats utilized by a substantial portion of the ICL methods discussed in this survey are developed based on the OpenAI regulation. 
Furthermore, most of these ICL methods incorporate zero-shot and few-shot prompting into their text-to-SQL frameworks.

As the mainstream research approach in LLM-based text-to-SQL, various well-designed methods within the ICL paradigm have been adopted to enhance LLM's SQL generation. 
In this section, we group these existing methods into five categories, $\mathbf{C}_{0:4}$, including $\mathbf{C}_{0}$-Vanilla Prompting, $\mathbf{C}_{1}$-Decomposition, $\mathbf{C}_{2}$-Prompt Optimization, $\mathbf{C}_{3}$-Reasoning Enhancement, and $\mathbf{C}_{4}$-Execution Refinement. 
The representative methods of each category are provided in Tab.~\ref{tab:icl-cat}. 
We will introduce each category accordingly and summarize their characteristics at the beginning of each corresponding paragraph.

\subsubsection{\textbf{$\mathbf{C}_{0}$-Vanilla Prompting}}

In our survey, we identified a category of prompting strategies that do not propose a well-designed ICL framework. 
These strategies are based solely on zero-shot and few-shot prompting. 
Despite their simplicity, they offer valuable insights and findings on the ICL paradigm in LLM-based text-to-SQL. 
We refer to these strategies as vanilla prompting strategies and further divide their studies into zero-shot and few-shot studies.

\begin{table}[!t]
\caption{A summary of representative methods used in the ICL paradigm for LLM-based text-to-SQL is provided. The complete table, along with a detailed categorization of existing well-designed methods denoted as $\mathbf{C}_{1:4}$, is available in Tab.~\ref{tab:icl-body}. \textsuperscript{*}Note that SimpleDDL is a prompting style that provides only table and column names~\cite{zhang2024benchmarking}, rather than a tailored method.}
\label{tab:icl-cat}
\centering
\begin{tabular}{lcc}
\toprule
Methods & Adopted by & Applied LLMs \\ 
\midrule
$\mathbf{C}_{0}$-Vanilla Prompting & SimpleDDL~\cite{zhang2024benchmarking}\textsuperscript{*} & SQLCoder \\
\midrule
$\mathbf{C}_{1}$-Decomposition & DIN-SQL~\cite{pourreza2023dinsql} & GPT-4 \\
$\mathbf{C}_{2}$-Prompt Optimization & DAIL-SQL~\cite{gao2023dailsql} & GPT-4 \\
$\mathbf{C}_{3}$-Reasoning Enhancement & ACT-SQL~\cite{zhang2023actsql} & GPT-4 \\
$\mathbf{C}_{4}$-Execution Refinement & LEVER~\cite{ni2023lever} & Codex \\
\bottomrule
\end{tabular}
\end{table}

\begin{table*}[!t]
    \caption{Well-designed methods in the ICL paradigm for LLM-based text-to-SQL are ordered by release time. The methods are grouped into four categories based on their implementation perspective: \textbf{$\mathbf{C}_{1}$-Decomposition}, \textbf{$\mathbf{C}_{2}$-Prompt Optimization}, \textbf{$\mathbf{C}_{3}$-Reasoning Enhancement}, and \textbf{$\mathbf{C}_{4}$-Execution Refinement}. Methods that fall into multiple categories will be introduced accordingly.
    The dataset column consists of six text-to-SQL datasets, which can be clarified by names: Spider~\cite{yu2018Spider}, BIRD~\cite{li2023BIRD}, Spider-DK~\cite{gan2021Spider-DK} Spider-SYN~\cite{gan2021Spider-SYN}, Spider-Realistic~\cite{deng2021Spider-Realistic}, and Spider 2.0~\cite{lei2025spider2}. We use the reference ID in the table due to column space limitations.
    \textsuperscript{*}There are multiple applied LLMs in the corresponding method; the model with representative performance is presented.
    \textsuperscript{\dag}CoT method is introduced in multiple venues: NeurIPS'23~\cite{li2023BIRD}, EMNLP'23~\cite{tai2023exploring}, VLDB'24~\cite{gao2023dailsql}, arXiv'24~\cite{zhang2024sgusql}.}
  \label{tab:icl-body}
  \centering
  \resizebox{0.99\textwidth}{!}{
  \begin{tabular}{@{}l|ccc|cccc|cc@{}}
  \toprule
    \textbf{Methods} & \textbf{Applied LLMs} & \textbf{Dataset} & \textbf{Metrics} & \textbf{$\mathbf{C}_{1}$} & \textbf{$\mathbf{C}_{2}$} & \textbf{$\mathbf{C}_{3}$} & \textbf{$\mathbf{C}_{4}$} & \textbf{Release Time} & \textbf{Publication Venue} \\
    \midrule
    MBR-Exec~\cite{shi2022natural} & Codex & \cite{yu2018Spider} & EX & & & & \checkmark & Apr-2022 & EMNLP'22 \\
    Coder-Reviewer~\cite{zhang2022coder} & Codex & \cite{yu2018Spider} & EX & \checkmark & & & \checkmark & Nov-2022 & ICML'23 \\
    LEVER~\cite{ni2023lever} & Codex & \cite{yu2018Spider} & EX & & & & \checkmark & Feb-2023 & ICML'23 \\
    Self-Debugging~\cite{chen2024selfdebugging} & StarCoder\textsuperscript{*} & \cite{yu2018Spider} & EX & & & & \checkmark & Apr-2023 & ICLR'24 \\
    DESEM+P~\cite{guo2023desem} & ChatGPT & \cite{yu2018Spider,gan2021Spider-SYN} & EX & & \checkmark & & \checkmark & Apr-2023 & PRICAI'23 \\
    DIN-SQL~\cite{pourreza2023dinsql} & GPT-4\textsuperscript{*} & \cite{li2023BIRD,yu2018Spider,lei2025spider2} & EX, EM, VES & \checkmark & & & \checkmark & Apr-2023 & NeurIPS'23 \\
    CoT~\cite{tai2023exploring,li2023BIRD, gao2023dailsql,zhang2024sgusql} & GPT-4 & \cite{yu2018Spider,deng2021Spider-Realistic,li2023BIRD} & EX, VES & & & \checkmark &  & May-2023 & Multiple Venues\textsuperscript{\dag} \\
    StructGPT~\cite{jiang2023structgpt} & ChatGPT\textsuperscript{*} & \cite{yu2018Spider,gan2021Spider-SYN,deng2021Spider-Realistic} & EX & & \checkmark & & & May-2023 & EMNLP'23 \\
    SD+SA+Voting~\cite{nan2023enhancing} & ChatGPT\textsuperscript{*} & \cite{yu2018Spider,gan2021Spider-SYN,deng2021Spider-Realistic} & EX & & \checkmark & & \checkmark & May-2023 & EMNLP'23 Findings \\
    QDecomp~\cite{tai2023exploring} & Codex & \cite{yu2018Spider,deng2021Spider-Realistic} & EX & \checkmark & & \checkmark & & May-2023 & EMNLP'23 \\
    Least-to-Most~\cite{tai2023exploring} & Codex & \cite{yu2018Spider} & EX & & & \checkmark & & May-2023 & EMNLP'23 \\
    SQL-PaLM~\cite{sun2023sqlpalm} & PaLM-2 & \cite{yu2018Spider} & EX & & & \checkmark & \checkmark & May-2023 & TMLR'24 \\
    RAG+SP\&DRC~\cite{guo2023ragsql} & ChatGPT & \cite{yu2018Spider} & EX & & \checkmark & & \checkmark & Jul-2023 & ICONIP'23 \\
    C3~\cite{dong2023c3} & ChatGPT & \cite{yu2018Spider} & EX & \checkmark & \checkmark & & \checkmark & Jul-2023 & arXiv'23 \\
    DAIL-SQL~\cite{gao2023dailsql} & GPT-4\textsuperscript{*} & \cite{yu2018Spider,li2023BIRD,deng2021Spider-Realistic,lei2025spider2} & EX, EM, VES & & \checkmark & \checkmark & & Aug-2023 & VLDB'24 \\
    ODIS~\cite{chang2023selective} & Codex\textsuperscript{*} & \cite{yu2018Spider} & EX & & \checkmark & & & Oct-2023 & EMNLP'23 Findings \\
    ACT-SQL~\cite{zhang2023actsql} & GPT-4\textsuperscript{*} & \cite{yu2018Spider,gan2021Spider-SYN} & EX, EM & & \checkmark & \checkmark & & Oct-2023 & EMNLP'23 Findings \\
    MAC-SQL~\cite{wang2024macsql} & GPT-4\textsuperscript{*} & \cite{yu2018Spider,li2023BIRD} & EX, EM, VES & \checkmark & & & \checkmark & Dec-2023 & COLING'25 \\
    DEA-SQL~\cite{xie2024deasql} & GPT-4 & \cite{yu2018Spider} & EX & \checkmark & & & & Feb-2024 & ACL'24 Findings \\
    FUSED~\cite{wang2024fused} & ChatGPT\textsuperscript{*} & \cite{yu2018Spider} & EX & & \checkmark & & & Feb-2024 & EMNLP'24 Findings \\
    DELLM~\cite{hong2024knowledgetosql} & GPT-4\textsuperscript{*} & \cite{yu2018Spider,li2023BIRD} & EX, VES & & \checkmark & & \checkmark & Feb-2024 & ACL'24 Findings \\
    SGU-SQL~\cite{zhang2024sgusql} & GPT-4\textsuperscript{*} & \cite{yu2018Spider,li2023BIRD} & EX, EM & \checkmark & & & & Feb-2024 & ICML'25 \\
    PoT~\cite{xia2024sqlcraft} & GPT-4\textsuperscript{*} & \cite{yu2018Spider,li2023BIRD} & EX & & & \checkmark & & Feb-2024 & arXiv'24 \\
    SQL-CRAFT~\cite{xia2024sqlcraft} & GPT-4\textsuperscript{*} & \cite{yu2018Spider,li2023BIRD} & EX & & & \checkmark & \checkmark & Feb-2024 & arXiv'24 \\
    FUXI~\cite{gu2024fuxi} & GPT-4\textsuperscript{*} & \cite{li2023BIRD} & EX & & & \checkmark & \checkmark & Feb-2024 & EMNLP'24 \\
    MetaSQL~\cite{fan2024metasql} & GPT-4\textsuperscript{*} & \cite{yu2018Spider} & EX, EM & \checkmark & & & & Feb-2024 & ICDE'24 \\
    $\textit{\textbf{R}}^3$~\cite{xia2024r3} & GPT-4\textsuperscript{*} & \cite{yu2018Spider,li2023BIRD} & EX & \checkmark & & & \checkmark & Feb-2024 & ACL'25 Workshop \\
    PET-SQL~\cite{li2024petsql} & GPT-4 & \cite{yu2018Spider} & EX & \checkmark & & & \checkmark & Mar-2024 & arXiv'24 \\
    PURPLE~\cite{ren2024purple} & GPT-4\textsuperscript{*} & \cite{yu2018Spider,gan2021Spider-SYN,deng2021Spider-Realistic} & EX, EM & \checkmark & & & \checkmark & Mar-2024 & ICDE'24 \\
    Dubo-SQL~\cite{thorpe2024dubo} & ChatGPT\textsuperscript{*} & \cite{li2023BIRD} & EX & & \checkmark & & \checkmark & Apr-2024 & arXiv'24 \\
    TA-SQL~\cite{qu2024generation} & GPT-4\textsuperscript{*} & \cite{yu2018Spider,li2023BIRD,gan2021Spider-DK,deng2021Spider-Realistic} & EX & \checkmark & \checkmark & & & Apr-2024 & ACL'24 Findings \\
    MCS-SQL~\cite{lee2025mcs} & GPT-4 & \cite{yu2018Spider,li2023BIRD} & EX, VES & \checkmark & \checkmark & & \checkmark & May-2024 & COLING'25 \\
    CHESS~\cite{talaei2024chess} & GPT-4o\textsuperscript{*} & \cite{yu2018Spider,li2023BIRD,lei2025spider2} & EX, VES & \checkmark & & & \checkmark & May-2024 & arXiv'24 \\
    SuperSQL~\cite{li2024dawn} & GPT-4\textsuperscript{*} & \cite{yu2018Spider,li2023BIRD} & EX, EM, VES & \checkmark & \checkmark & \checkmark & & Jun-2024 & VLDB'24 \\
    Distillery~\cite{maamari2024the} & GPT-4o-Mini\textsuperscript{*} & \cite{li2023BIRD} & EX & \checkmark & \checkmark & & \checkmark & Aug-2024 & NeurIP'24 Workshop \\
    E-SQL~\cite{caferouglu2024esql} & GPT-4o\textsuperscript{*} & \cite{yu2018Spider,li2023BIRD} & EX, VES &  & & & \checkmark & Sep-2024 & arXiv'24 \\
    CHASE-SQL~\cite{pourreza2025chasesql} & Gemini-1.5\textsuperscript{*} & \cite{yu2018Spider,li2023BIRD} & EX, VES & & & \checkmark & \checkmark & Oct-2024 & ICLR'25 \\
    RSL-SQL~\cite{cao2024rsl} & DeepSeek-V3\textsuperscript{*} & \cite{yu2018Spider,li2023BIRD} & EX, VES & \checkmark & \checkmark & & \checkmark & Nov-2024 & arXiv'24 \\
    Spider-Agent~\cite{lei2025spider2} & o1-preview\textsuperscript{*} & \cite{lei2025spider2} & EX & \checkmark & & \checkmark & \checkmark & Nov-2024 & ICLR'25 \\
    ICRL~\cite{toteja2025context} & ChatGPT & \cite{yu2018Spider} & EX & & \checkmark & & & Jan-2025 & COLING'25 \\
    Gen-SQL~\cite{shi2025gen} & LLaMA-3\textsuperscript{*} & \cite{yu2018Spider,li2023BIRD} & EX, EM, VES & \checkmark & \checkmark & & & Jan-2025 & COLING'25 \\
    SAFE-SQL~\cite{lee2025safe} & GPT-4o-Mini\textsuperscript{*} & \cite{yu2018Spider} & EX, EM & & \checkmark & & & Feb-2025 & arXiv'25 \\
    ReFoRCE~\cite{deng2025reforce} & o1-preview\textsuperscript{*} & \cite{lei2025spider2} & EX & \checkmark & \checkmark & \checkmark & \checkmark & Feb-2025 & ICLR'25 Workshop \\
    LinkAlign~\cite{wang2025linkalign} & DeepSeek-R1\textsuperscript{*} & \cite{li2023BIRD,lei2025spider2} & EX, EM & \checkmark & \checkmark & & & Mar-2025 & arXiv'25 \\
    \bottomrule
  \end{tabular}}
\end{table*}

\noindent {\bf{1. Zero-shot studies:}}
The zero-shot-based ICL strategies primarily focus on studying the influence of prompt representation styles and the baseline performance of various LLMs for text-to-SQL~\cite{rajkumar2022evaluating,liu2023comprehensive,zhang2024benchmarking}.
As an empirical evaluation, \cite{rajkumar2022evaluating} evaluates the baseline text-to-SQL capabilities of various early-developed LLMs~\cite{raffel2020exploring,brown2020gpt3,chen2021evaluating} and their performance across different prompting styles. 
The experimental results indicate that prompt design is critical for performance. 
Through error analysis, this study \cite{rajkumar2022evaluating} further suggests that including more database content can harm the overall accuracy of text-to-SQL.
Since the emergence of ChatGPT~\cite{brown2020gpt3} with its impressive capabilities in conversational scenarios and code generation~\cite{ray2023chatgpt}, \cite{liu2023comprehensive} evaluates its performance on Spider~\cite{yu2018Spider} and its variant~\cite{gan2021Spider-SYN,gan2021Spider-DK,deng2021Spider-Realistic} datasets. 
Under zero-shot settings, the results demonstrate that ChatGPT has promising text-to-SQL performance compared to state-of-the-art PLM-based systems. 
Aside from the proprietary models discussed above, similar evaluations indicate that open-source LLMs can also perform zero-shot text-to-SQL tasks~\cite{xue2024dbgpt,xu2024symbolllm,zhang2024benchmarking}, particularly for the code-specific models~\cite{chen2024selfdebugging,zhang2024benchmarking}.

Primary keys and foreign keys convey continuous knowledge across different tables, making them an effective prompting style in zero-shot scenarios.
\cite{zhang2023actsql} studied their impact by incorporating these keys into various prompt representation styles with different database content to analyze performance outcomes. 
A benchmark evaluation~\cite{gao2023dailsql} also assessed the influence of foreign keys using five distinct prompt representation styles. Each style can be viewed as a permutation and combination of instruction, rule implication, and foreign key elements. 
In addition to using foreign keys, this study explores zero-shot prompting combined with the rule implication \textit{``no explanation''} to produce concise outputs. With the help of annotated external knowledge from human experts, the evaluation in the BIRD benchmark~\cite{li2023BIRD} significantly improves by incorporating the provided annotated oracle knowledge. 

In terms of zero-shot prompting optimization, \cite{zhang2024benchmarking} highlights the challenge of designing effective prompt templates for LLMs. 
Previous simple prompt constructions often lack structural uniformity, complicating the identification of specific elements in a prompt template that influence LLM performance. 
This challenge is addressed by investigating a more unified set of prompt templates that include various prefixes, infixes, and postfixes.
To ensure fair comparability, \cite{chang2023how} explores effective prompt construction for LLMs by studying different styles of prompt representation and highlights their findings on the crucial role of table structure, content, prompt length, and optimal representations in effective prompting.

\noindent {\bf{2. Few-shot studies:}} The study of few-shot prompting strategies focuses on the number and selection of few-shot instances. 
Empirical studies have evaluated few-shot prompting for text-to-SQL across multiple datasets using various LLMs~\cite{pourreza2023dinsql,zhang2024sgusql}, demonstrating solid improvements compared to zero-shot prompting. 
\cite{li2023BIRD} provides a detailed 1-shot example to trigger the text-to-SQL model for generating accurate SQL. 
\cite{xia2024sqlcraft} explores the effect of the number of few-shot examples. 
\cite{nan2023enhancing} focuses on sampling strategies by examining the similarity and diversity between different demonstrations, setting random sampling as the baseline, and evaluating various strategies and their combinations for comparison. 
Furthermore, beyond similarity-based selection, \cite{gao2023dailsql} evaluates masked question similarity selection and the upper limit of similarity approaches with various numbers of few-shot examples.

A study on difficulty-level sample selection~\cite{tai2023exploring} compared the performance of few-shot Codex~\cite{chen2021evaluating}, using random selection and difficulty-based selection for few-shot instances on difficulty-categorized datasets~\cite{yu2018Spider,deng2021Spider-Realistic}. 
Three difficulty-based selection strategies are devised based on samples selected by difficulty level. 
\cite{zhang2023actsql} utilizes a hybrid strategy for selecting samples, combining static examples and similarity-based dynamic examples for few-shot prompting. 
In their settings, they also evaluate the impact of different input schema styles and various numbers of static and dynamic exemplars.

The impact of cross-domain examples is also being studied~\cite{chang2023selective}. 
When incorporating in- and out-of-domain examples in different numbers, the in-domain demonstrations outperform zero-shot and out-of-domain examples, with performance improving as the number of examples increases. 
To explore the detailed construction of prompts, \cite{sun2023sqlpalm} compares concise and verbose prompt design strategies. 
The former style separates the schema, column names, primary keys, and foreign keys, while the latter organizes them as natural language descriptions. 

\subsubsection{\textbf{$\mathbf{C}_{1}$-Decomposition}}\label{sec:decompose}
Decomposition-based methods reduce text-to-SQL complexity by dividing tasks or questions into manageable components, employing two main strategies:

\noindent {\bf{1. Multi-module collaboration:}}
This strategy breaks the task into specialized stages (e.g., schema linking, SQL generation, and SQL refinement) handled by distinct modules or agents. 
C3~\cite{dong2023c3} employs a three-module pipeline involving schema linking via clear prompting, calibration bias prompting for SQL guidance, and consistency-based voting for final selection. 
Similarly, MAC-SQL~\cite{wang2024macsql} and CHESS~\cite{talaei2024chess} use multi-agent frameworks where dedicated agents handle schema selection, decomposition, and refinement. 
Workflow-centric methods, like Distillery~\cite{maamari2024the} and DEA-SQL~\cite{xie2024deasql}, decompose the task into retrieval, generation, and correction stages.
Iterative refinement and structural decomposition are also prevalent. 
DIN-SQL~\cite{pourreza2023dinsql} integrates a four-stage pipeline, including schema linking, classification, SQL generation, and self-correction. 
Coder-Reviewer~\cite{zhang2022coder} separates generation and validation into distinct models. 
PET-SQL~\cite{li2024petsql} refines outputs through multi-stage prompting, while MCS-SQL~\cite{lee2025mcs} uses execution-guided selection as the final module after multiple SQL generation. 
With the increasing complexity of recently proposed benchmarks, agentic frameworks like ReFoRCE~\cite{deng2025reforce}, LinkAlign~\cite{wang2025linkalign} and Spider-Agent~\cite{lei2025spider2} leveraged well-designed step-by-step action plans for LLMs to systematically reason toward the final SQL.

\noindent {\bf{2. Question decomposition:}}
Complexity is reduced by splitting user questions into intermediate sub-questions via question decomposition. 
QDecomp~\cite{tai2023exploring} uses Least-to-Most prompting, breaking complex questions into step-by-step sub-questions to facilitate SQL generation. 
SGU-SQL~\cite{zhang2024sgusql} employs graph-based structure linking and meta-operators for decomposing grammar-aware sub-questions. 
Metadata-driven approaches, such as MetaSQL~\cite{fan2024metasql} and TA-SQL~\cite{qu2024generation}, further decompose questions using metadata conditioning and task-aligned schema linking. 
Meanwhile, frameworks like DIN-SQL~\cite{pourreza2023dinsql} and MAC-SQL~\cite{wang2024macsql} use question decomposition as auxiliary modules but primarily focus on module-level collaboration.

\subsubsection{\textbf{$\mathbf{C}_{2}$-Prompt Optimization}}\label{sec:prompt-opt}
Prompt optimization methods for text-to-SQL focus on enhancing prompt engineering quality. 
These methods are categorized into three key strategies:

\noindent {\bf{1. Advanced few-shot sampling:}}
This strategy optimizes example selection through semantic similarity, diversity, or domain alignment. 
DESEM~\cite{guo2023desem} and retrieval-augmented frameworks~\cite{guo2023ragsql,thorpe2024dubo,maamari2024the} de-semanticize questions to retrieve examples with matching intent skeletons or vector embeddings. 
DAIL-SQL~\cite{gao2023dailsql} balances quality and quantity by masking domain-specific tokens and ranking examples based on the Euclidean distance between questions and SQL queries, while SAFE-SQL~\cite{lee2025safe} combines skeleton masking with self-augmentation. 
ODIS~\cite{chang2023selective} hybridizes out-of-domain and synthetic in-domain examples for robustness, and FUSED~\cite{wang2024fused} enhances diversity through iterative demonstration synthesis. 
Dynamic selection is explored in ACT-SQL~\cite{zhang2023actsql} and MCS-SQL~\cite{lee2025mcs}, which adaptively retrieve examples based on defined similarity scores or masked similarity scores.

\noindent {\bf{2. Schema augmentation:}}
These methods refine schema representations to reduce redundancy and improve relevance. 
C3~\cite{dong2023c3} distills question-relevant schemas by removing unrelated tables or columns, while RSL-SQL~\cite{cao2024rsl} simplifies schemas using NL descriptions. 
ReFoRCE~\cite{deng2025reforce} employs table compression and column exploration to manage long-context schema complexity. 
Gen-SQL~\cite{shi2025gen} bridges NL-SQL gaps by generating pseudo-schemas that align with user questions. 
QDecomp~\cite{tai2023exploring} and SD+SA+Voting~\cite{nan2023enhancing} further augment prompts with schema-aware semantic and structural hints. 
For more complex databases, LinkAlign~\cite{wang2025linkalign} introduces a schema-based question rewriting strategy, combined with a retrieve-then-filter schema selection mechanism for accurate linking.

\noindent {\bf{3. External knowledge integration:}}
This group enhances prompts with domain-specific knowledge or feedback mechanisms. 
Knowledge-to-SQL~\cite{hong2024knowledgetosql} trains a DELLM to inject external knowledge about question-database relationships, while ICRL~\cite{toteja2025context} integrates reinforcement learning with schema-graph-derived knowledge bases. 
MAC-SQL~\cite{wang2024macsql} and DIN-SQL~\cite{pourreza2023dinsql} implicitly utilize decomposition patterns as procedural knowledge to guide generation. 
C3~\cite{dong2023c3} and SAFE-SQL~\cite{lee2025safe} incorporate calibration bias and self-augmented feedback loops, respectively, to iteratively refine outputs.

\subsubsection{\textbf{$\mathbf{C}_{3}$-Reasoning Enhancement}}\label{sec:reasoning}
Reasoning enhancement methods\footnote{Unlike decomposition paradigms, these methods retain single-turn generation while enhancing intrinsic reasoning through structured prompting.} improve LLMs' ability to handle complex SQL through structured reasoning and consistency mechanisms. 
These approaches focus on three crucial strategies:

\noindent {\bf{1. Chain-of-thought (CoT)-based reasoning:}}
CoT-based methods guide LLMs in generating intermediate reasoning steps before predicting SQLs. 
While standard CoT prompts (e.g., ``Let’s think step by step'') show limited effectiveness in text-to-SQL~\cite{gao2023dailsql,li2023BIRD}, adaptations help bridge this gap. 
ACT-SQL~\cite{zhang2023actsql} automates CoT example generation by linking question slices to database columns via similarity matching. 
QDecomp~\cite{tai2023exploring} restructures SQL clauses into logical steps ordered by execution flow, reducing error propagation compared to generic CoT. 
CHASE-SQL~\cite{pourreza2025chasesql} extends this with multi-path reasoning for candidate generation using ``divide and conquer'' CoT strategy.

\noindent {\bf{2. Consistency-driven reasoning:}}
These methods improve reliability through candidate diversity and agreement metrics before execution. 
Self-consistency~\cite{wang2023selfconsistency} is employed in frameworks like C3~\cite{dong2023c3}, DAIL-SQL~\cite{gao2023dailsql}, and SuperSQL~\cite{li2024dawn}, where majority voting selects the most frequent SQL candidate based on syntax or structural similarity. 
SD+SA+Voting~\cite{nan2023enhancing} refines this by filtering candidates using schema-aware rules (e.g., valid joins, clause ordering) before voting. 
Unlike execution-dependent methods, these strategies prioritize logical consistency and syntactic validity over runtime feedback, making them lightweight but less robust to semantic errors.

\noindent {\bf{3. Tool-augmented and agentic reasoning:}}
Hybrid techniques combine SQL generation with external computation or tools. 
SQL-CRAFT~\cite{xia2024sqlcraft} employs Program-of-Thoughts~\cite{chen2023program} (PoT) prompting, requiring LLMs to generate Python code alongside SQL for arithmetic-heavy queries, improving performance on datasets like BIRD~\cite{li2023BIRD}. 
FUXI~\cite{gu2024fuxi} integrates dedicated tools for schema linking and syntax checking, dynamically invoking them during generation. 
These approaches differ from pure prompting strategies by explicitly bridging LLM reasoning with external symbolic systems.
Also, recent frameworks such as ReFoRCE~\cite{deng2025reforce} and Spider-Agent~\cite{lei2025spider2} adopt step-by-step agentic reasoning, where LLMs act as specialized agents that iteratively plan actions to handle complex database scenarios.

\subsubsection{\textbf{$\mathbf{C}_{4}$-Execution Refinement}}\label{sec:execution}
Execution refinement methods improve SQL accuracy by leveraging database feedback through regeneration or preferred SQL selection. 
These approaches are categorized into two paradigms:

\noindent {\bf{1. Feedback-driven regeneration:}}
These methods iteratively refine SQL queries using execution errors or intermediate results. 
Self-Debugging~\cite{chen2024selfdebugging} enables LLMs to autonomously correct errors via natural language explanations of failed executions, while DESEM~\cite{guo2023desem} uses error-type-specific prompts with termination criteria to prevent infinite loops. 
Multi-agent frameworks like MAC-SQL~\cite{wang2024macsql} and CHASE-SQL~\cite{pourreza2025chasesql} utilize agents (e.g., Refiner, Query Fixer) to diagnose SQLite errors and regenerate SQL fixes. 
SQL-CRAFT~\cite{xia2024sqlcraft} introduces adaptive control to balance correction depth, preventing over-refinement. 
Schema-focused methods like RSL-SQL~\cite{cao2024rsl} and DIN-SQL~\cite{pourreza2023dinsql} use generic or gentle prompts to help LLMs identify schema misalignments during regeneration. 
Knowledge-to-SQL~\cite{hong2024knowledgetosql} optimizes its DELLM model through preference learning from execution feedback, while FUXI~\cite{gu2024fuxi} integrates error feedback into tool-based reasoning. 
E-SQL~\cite{caferouglu2024esql} enhances questions with execution-aware candidate predictions.
Agentic frameworks~\cite{lei2025spider2} also incorporate feedback-driven regeneration, where specialized agents iteratively analyze execution feedback to plan corrective actions to refine SQLs.

\noindent {\bf{2. Execution-guided selection:}}
This strategy leverages database execution results to select or rank candidates. 
ReFoRCE~\cite{deng2025reforce} and PET-SQL~\cite{li2024petsql} prune invalid SQL queries that fail execution, then apply bias-aware voting (e.g., difficulty-weighted or efficiency-prioritized as in MCS-SQL~\cite{lee2025mcs}). 
Probabilistic verification distinguishes methods: MRC-EXEC~\cite{shi2022natural} minimizes Bayes risk by comparing execution outputs to expected results, and LEVER~\cite{ni2023lever} trains a verifier to predict correctness likelihoods from execution traces. 
Hybrid approaches like the retrieval-augmented framework~\cite{guo2023ragsql} use execution feedback for iterative SQL revision through semantic gap analysis, bridging generation and selection.

\subsection{Fine-tuning}\label{sec:ft}
\begin{table*}[!t]
  \caption{Well-designed methods used in the FT paradigm for LLM-based text-to-SQL. The methods in each category are ordered by release time. \textsuperscript{*}The methods are utilized in multiple open-source LLMs; we select a representative model to present.}
  \label{tab:ft-body}
  \centering
  \resizebox{\textwidth}{!}{
  \begin{tabular}{@{}@{}@{}l|l|ccccc|cc@{}@{}@{}}
    \toprule
    \textbf{Category} & \textbf{Adopted by} & \textbf{Applied LLMs} & \textbf{Dataset} & \textbf{EX} & \textbf{EM} & \textbf{VES} & \textbf{Release Time} & \textbf{Publication Venue}\\
    \midrule

    \multirow{1}{*}{Enhanced Architecture}
    &
    CLLMs~\cite{kou2024cllms}
    & DeepSeek\textsuperscript{*} & \cite{yu2018Spider} & \checkmark & & & Mar-2024 & ICML'24 \\
    \midrule
    
    \multirow{1}{*}{Pre-training} & CodeS~\cite{li2024codes}
    & StarCoder & \cite{yu2018Spider,li2023BIRD} & \checkmark & & \checkmark & Feb-2024 & SIGMOD'24 \\
    \midrule
    
    \multirow{8}{*}{Data Augmentation}
    & 
    DAIL-SQL~\cite{gao2023dailsql}
    & LLaMA\textsuperscript{*} & \cite{yu2018Spider,deng2021Spider-Realistic} & \checkmark & \checkmark & & Aug-2023 & VLDB'24 \\
    & 
    Symbol-LLM~\cite{xu2024symbolllm} 
    & CodeLLaMA & \cite{yu2018Spider} & & \checkmark & & Nov-2023 & ACL'24 \\
    & 
    CodeS~\cite{li2024codes}
    & StarCoder & \cite{yu2018Spider,li2023BIRD,lei2025spider2} & \checkmark & & \checkmark & Feb-2024 & SIGMOD'24 \\
    & 
    StructLM~\cite{zhuang2024structlm}
    & CodeLLaMA & \cite{yu2018Spider} & & \checkmark & & Feb-2024 & COLM'24 \\
    &
    Dubo-SQL~\cite{thorpe2024dubo}
    & ChatGPT & \cite{li2023BIRD} & \checkmark & & & Apr-2024 & arXiv'24 \\
    &
    Distillery~\cite{maamari2024the}
    & LLaMA-3.1\textsuperscript{*} & \cite{li2023BIRD} & \checkmark & & & Aug-2024 & NeurIPS'24 Workshop \\
    & 
    XiYan-SQL~\cite{gao2024xiyan}
    & Qwen2.5-Coder & \cite{yu2018Spider,li2023BIRD} & \checkmark & & & Nov-2024 & arXiv'24 \\
    & 
    SHARE~\cite{qu2025share}
    & LLaMA-3.1\textsuperscript{*} & \cite{yu2018Spider,li2023BIRD,gan2021Spider-DK,deng2021Spider-Realistic} & \checkmark & & & Jun-2025 & ACL'25 \\

    \midrule
    \multirow{6}{*}{Multi-task Tuning}
    & 
    SQL-LLaMA~\cite{wang2024macsql}
    & CodeLLaMA & \cite{yu2018Spider,li2023BIRD} & \checkmark & & \checkmark & Dec-2023 & COLING'25 \\
    & 
    DTS-SQL~\cite{pourreza2024dtssql}
    & Mistral\textsuperscript{*} & \cite{yu2018Spider,gan2021Spider-SYN} & \checkmark & \checkmark & & Feb-2024 & EMNLP'24 Findings \\
    &
    KaSLA~\cite{yuan2025knapsack}
    & DeepSeek-Coder\textsuperscript{*} & \cite{yu2018Spider,li2023BIRD} & \checkmark & & \checkmark & Oct-2024 & arXiv'24 \\
    &
    MSc-SQL~\cite{gorti2024mscsql}
    & Gemma-2 & \cite{yu2018Spider,li2023BIRD} & \checkmark & & & Oct-2024 & NAACL'25 \\
    & 
    XiYan-SQL~\cite{gao2024xiyan}
    & Qwen2.5-Coder & \cite{yu2018Spider,li2023BIRD} & \checkmark & & & Nov-2024 & arXiv'24 \\
    & 
    ROUTE~\cite{qin2025route}
    & Qwen2.5\textsuperscript{*} & \cite{yu2018Spider,li2023BIRD,gan2021Spider-SYN,gan2021Spider-DK,deng2021Spider-Realistic} & \checkmark & & \checkmark & Feb-2025 & ICLR'25 \\
    & 
    SHARE~\cite{qu2025share}
    & LLaMA-3.1\textsuperscript{*} & \cite{yu2018Spider,li2023BIRD,gan2021Spider-DK,deng2021Spider-Realistic} & \checkmark & & & Jun-2025 & ACL'25 \\
    \bottomrule
  \end{tabular}}
\end{table*}
Supervised Fine-tuning (SFT) is the widely-used approach within the FT paradigm for training LLMs~\cite{zhao2023survey,zheng2024llamafactory}. For open-source LLMs, like LLaMA~\cite{touvron2023llama2} and Qwen~\cite{bai2023qwen} series, the most straightforward way to enable the model to adapt to a specific domain is to apply SFT over collected domain-specific data. The SFT phase typically serves as an initial step in a well-designed training pipeline~\cite{rafailo2023direct,ouyang2022training}, including LLM-based text-to-SQL training.
Consider the prompt defined in Eq.~\ref{eq:zero-shot} and \ref{eq:few-shot} as $P$, and let $Y = [y_1, y_2, \dots, y_T]$ represent the target ground-truth SQL query. The probability $\text{Pr}$ for an LLM $\pi$ to generate $Y$ given $P$ is formally modeled as:
\begin{equation}
    \text{Pr}_\pi\left(Y \mid P\right) = \prod_{t=1}^{T} \text{Pr}_\pi\left(y_t \mid P, y_{<t}\right).
\end{equation}
Contrary to the LLM in the ICL paradigm, the parameters of $\pi$ are tuned. The SFT process minimizes the cross-entropy loss over the training dataset $\mathcal{D}$ for $\pi$:
\begin{equation}
\mathcal{L}_{\text{SFT}} = -\sum_{\left(P,\,Y\right) \in \mathcal{D}} \sum_{t=1}^{T} \log \text{Pr}_\pi\left(y_t \mid P, y_{<t}\right).
\end{equation}
As a vanilla fine-tuning approach for LLM-based text-to-SQL, SFT is widely adopted in text-to-SQL research across various open-source LLMs~\cite{li2024codes,zhang2024benchmarking,gao2023dailsql}.
FT paradigms are often more foundational in LLM-based text-to-SQL than ICL, which is the mainstream approach in the community. 
This is largely due to the comparably limited capabilities of open-source LLMs versus proprietary models in understanding complex database content. 
For the existing well-designed methods in the FT paradigm, we group these methods into different categories based on their training mechanisms, as shown in Tab.~\ref{tab:ft-body}.

\subsubsection{\textbf{Enhanced Architecture}}
The widely-used Generative Pre-trained Transformer (GPT) framework~\cite{radford2018improving} employs a decoder-only transformer architecture and conventional auto-regressive decoding for text generation. 
With advancements in the GPT framework, enhancements in LLM-based text-to-SQL involve modifications of the standard transformer backbone. These modifications incorporate tailored architecture to efficiently handle the structural and syntactic complexities of SQL.
In traditional text-to-SQL research, this tailored architecture has effectively enhanced PLM-based performance. 
Techniques include schema-aware encoding~\cite{li2023graphix}, optimized attention~\cite{gong2024graph}, and targeted decoding~\cite{cao2023heterogeneous}, often relying on graph structures.
Most recently, the speed of generating SQL queries by LLMs is notably slower than traditional PLM-based methods~\cite{li2023resdsql,yang2024harnessing}, posing a challenge to building high-efficiency local NLIDBs.

As a solution, CLLMs~\cite{kou2024cllms} are designed to address this challenge with specialized decoding, offering speedup efficiency for SQL generation and reducing high latency.

\subsubsection{\textbf{Pre-training}}
Pre-training is a fundamental phase of the complete training process, aimed at acquiring text generation capabilities through auto-regressive training on extensive data~\cite{erhan2010does}. 
Code-specific LLMs (e.g., CodeLLaMA~\cite{roziere2023codellama}, StarCoder~\cite{li2023starcoder}) are pre-trained on code data, with diverse programming languages enabling them to generate code per user instructions~\cite{chen2021evaluating, yufei2023aligning}. 
For SQL-specific pre-training, a key challenge is that SQL/database-related content makes up only a small part of the pre-training corpus. 
As a result, the open-source LLMs do not acquire a solid understanding of converting NL questions to SQL during their pre-training. 

PLM-based methods propose initial attempts in pre-training paradigms, such as STAR~\cite{cai2022star} and multitask approaches~\cite{giaquinto2023multitask}, which improve SQL generation by leveraging SQL-guided and multitask pre-training strategies. 
For the most recent LLM-based text-to-SQL, the pre-training phase of the CodeS~\cite{li2024codes} model consists of three stages of incremental pre-training. 
Starting from a code-specific LLM~\cite{li2023starcoder}, CodeS are incrementally pre-trained on a hybrid training corpus, including SQL-related data, NL-to-Code data, and NL-related data. 
The text-to-SQL understanding and performance are significantly improved.

\subsubsection{\textbf{Data Augmentation}}
During the fine-tuning process, the quality of training labels is the most straightforward factor affecting the model's performance~\cite{song2022learningnoise}. 
Training with low-quality or absent labels is \textit{``making bricks without straw.''} 
Using high-quality or augmented data for fine-tuning consistently outperforms the meticulously designed methods relying on low-quality or raw data~\cite{deng2022recent,zhao2023survey}. 
Data-augmented fine-tuning in text-to-SQL has made substantial progress by focusing on enhancing data quality and efficiency during the SFT process.

DAIL-SQL~\cite{gao2023dailsql} is designed as an ICL framework that uses a sampling strategy to enhance few-shot instances. 
Incorporating these sampled instances in the SFT process boosts the performance of open-source LLMs. 
Symbol-LLM~\cite{xu2024symbolllm} proposes injection and infusion stages for data-augmented instruction tuning. 
CodeS~\cite{li2024codes} augments the training data with bi-directional generation aided by ChatGPT. 
StructLM~\cite{zhuang2024structlm} trains on multiple structured knowledge tasks to improve overall capability.
Dubo-SQL~\cite{thorpe2024dubo} created concise prompt templates for fine-tuning, which reduces costs and simultaneously achieves record performance. 
Distillery~\cite{maamari2024the} employs an iterative failure-aware approach, selecting new training samples based on error analysis.
XiYan-SQL~\cite{gao2024xiyan} uses basic-syntax SFT and generation-enhanced SFT, employing hybrid augmented training data to fine-tune the SQL generator.
SHARE~\cite{qu2025share} designed a rule-based error perturbation to augment schema linking and final SQL inference based on the decomposed trajectories.

\subsubsection{\textbf{Multi-task Tuning}}
Dividing a complex task into multiple tasks or utilizing multiple models is a straightforward approach to tackle complexity, as outlined in Sec.~\ref{sec:decompose} on ICL paradigms. 
Proprietary models in ICL-based methods boast a large number of parameters, surpassing the scale of open-source models used in fine-tuning. 
These models are adept at performing assigned sub-tasks effectively, facilitated by mechanisms like few-shot learning~\cite{dong2023c3,wang2024macsql}. 
To replicate this success, it is essential to assign sub-tasks strategically to open-source models, such as generating external knowledge, schema linking, and execution-based refinement, with appropriate training.
This involves constructing specific datasets for fine-tuning, ultimately aiding in the final SQL generation.

SQL-LLaMA~\cite{wang2024macsql} is a fine-tuned LLM framework utilizing multi-agent collaboration.
DTS-SQL~\cite{pourreza2024dtssql} introduced a two-stage multi-task text-to-SQL fine-tuning framework, incorporating a schema-linking pre-generation task before the final SQL generation.
Similarly, KaSLA~\cite{yuan2025knapsack} employed a two-stage framework using knapsack optimization as a fine-tuning strategy for the schema linking task, and SHARE~\cite{qu2025share} adopted three sequential models for SQL correction.
MSc-SQL~\cite{gorti2024mscsql} trained separate models for schema linking, SQL generation, and critiquing the predicted SQL query.
The ROUTE~\cite{qin2025route} framework involved the SFT process over noise filtering, data synthesis, and continued writing, validating the effectiveness of the multi-task tuning pipeline through collaboration prompting.

\section{Summary}
This section contrasts traditional and LLM-based approaches in addressing text-to-SQL challenges, focusing on ambiguity resolution, schema linking, complex SQL generation, and cross-domain adaptation. Additionally, we analyze differences between different categories of LLM-based approaches, while highlighting their strengths and trade-offs.
\subsection{LLM-based Methods vs. Traditional Methods.}

\subsubsection{Linguistic Complexity and Ambiguity}

Traditional rule-based systems, like GRAPPA~\cite{yu2021grappa}, decompose questions via syntactic templates but fail to resolve ambiguous phrasing. Neural models like RyenSQL~\cite{choi2021ryansql} improve lexical matching but lack contextual reasoning. In contrast, LLM-based methods such as DAIL-SQL~\cite{gao2023dailsql} leverage chain-of-thought prompting to infer dependencies stepwise, while C3-SQL~\cite{dong2023c3} iteratively validates SQL drafts against paraphrased questions.
While traditional methods offer lightweight disambiguation, their reliance on limited training data restricts their ability to handle unseen phrasings. LLM-based methods address this with broader pretraining, but their computational cost grows quadratically with context length, making them impractical for low-resource edge devices. A promising direction is a two-stage pipeline: use traditional methods for initial ambiguity detection, then invoke LLMs only for resolving flagged ambiguities. This hybrid approach balances efficiency and robustness.

\subsubsection{Schema Understanding and Representation}

Early systems rank schema elements via string similarity or schema-agnostic embeddings, truncating complex relationships. LLM-based methods enhance schema linking through in-context learning and structured attention. For example, DIN-SQL~\cite{pourreza2023dinsql} serializes schema into natural language prompts with foreign-key hints, while KaSLA~\cite{yuan2025knapsack} mirrors human prioritization of schema elements, reducing the common issues of omissions and redundancies in schema linking.
While rule-based methods are interpretable, their rigidity limits adaptability. LLM-based methods excel at semantic linking but lack transparency: their attention weights may prioritize incorrect columns due to spurious correlations. To mitigate this, future systems could combine symbolic rules with LLM inference: first, apply rule-based filters to exclude irrelevant columns, then use LLMs to rank the remaining candidates. A hybrid approach that combines LLMs with domain-specific rules can offer a balanced trade-off between performance and resource consumption.

\subsubsection{Rare and Complex SQL Operations}

Handling operations like nested sub-queries or window functions requires labor-intensive template engineering. For instance, previous rule-based methods~\cite{li2014constructing} manually define 40+ SQL grammar rules to map question patterns, which could not be generalized to unseen formulations. Neural semantic parsers like PICARD~\cite{scholak2021picard} constrain decoding via transition-based incremental parsing, ensuring syntactically valid output by rejecting invalid tokens at each step. 
LLM-based methods generate rare SQLs by combining parametric knowledge from code-centric pretraining with few-shot demonstrations. 
For example, DAIL-SQL~\cite{gao2023dailsql} injects prompts with examples of window functions, enabling the model to generalize to unseen formulations. 
Traditional methods like PICARD~\cite{scholak2021picard} guarantee syntactic validity via grammar-constrained decoders but cannot innovate on rare and complex SQL operations. 
LLM-based methods provide novel solutions but may produce invalid SQL. 
The most current agentic reasoning~\cite{lei2025spider2,deng2025reforce} mitigates invalid outputs through multi-agent actions, but still suffers from error accumulation.
A synergistic approach would integrate LLM's creativity with traditional validation: first, generate candidate queries with an LLM, then apply PICARD-like grammars to reject invalid tokens during beam search. This mirrors the ``generate-then-verify'' paradigm in program synthesis, where LLMs propose solutions and symbolic checkers enforce constraints.

\subsubsection{Cross-Domain Generalization}


Rule-based systems require manual template adjustments per domain, while PLM-based models like S$^2$SQL~\cite{hui2022s} rely on domain-specific fine-tuning. LLM-based methods like CodeS~\cite{li2024codes} adapt to niche domains (e.g., biomedical terminology) via code-centric pretraining.
Traditional domain adaptation methods require moderate fine-tuning data to align latent spaces but provide precise control over domain-specific mappings. LLM-based methods enable few-shot adaptation but risk overgeneralization. A viable solution is using LLMs as meta-optimizers: train a small adapter network for LLM using domain-specific rules, allowing the LLM to dynamically adjust its linking strategy. This combines the efficiency of rule-based adaptation with the LLM’s capacity to learn cross-domain analogies.

\subsection{Comparison within LLM-based Methods}
\subsubsection{Comparison within ICL paradigm}
The in-context learning (ICL) paradigm's categories exhibit complementary strengths and limitations. 
Decomposition ($\mathbf{C}_{1}$, Sec.~\ref{sec:decompose}) is effective at managing complex queries through task modularization but introduces cascading error risks from multi-stage pipelines. 
Prompt optimization ($\mathbf{C}_{2}$, Sec.~\ref{sec:prompt-opt}) enhances input quality through semantic alignment but faces challenges with inherent reasoning limitations. 
Reasoning enhancement ($\mathbf{C}_{3}$, Sec.~\ref{sec:reasoning}) improves logical coherence through structured thinking but significantly increases computational overhead.
Execution refinement ($\mathbf{C}_{4}$, Sec.~\ref{sec:execution}) ensures runtime validity through feedback loops but relies heavily on database accessibility. 
Recent methods like SuperSQL~\cite{li2024dawn}, Spider-Agent~\cite{lei2025spider2}, and ReFoRCE~\cite{deng2025reforce} illustrate a trend toward hybrid integration-combining decomposition's structural rigor with prompt optimization's efficiency, while layering reasoning enhancement and execution consistency, which are implemented through well-designed agentic reasoning actions.
This convergence suggests that future ICL methods will increasingly unify cross-category techniques into cohesive frameworks, balancing accuracy, efficiency, and robustness through systematic collaboration.

\subsubsection{Comparison within FT paradigm} 
Fine-tuning (FT) approaches highlight the trade-off between specialization and generalization. 
Enhanced architectures such as CLLMs~\cite{kou2024cllms} reduce latency, while pre-training approaches like CodeS~\cite{li2024codes} offer robust SQL foundations. Consequently, both approaches are constrained by their demands on computational resources and hardware devices.
Data augmentation (DAIL-SQL~\cite{gao2023dailsql}, XiYan-SQL~\cite{gao2024xiyan}) improves data efficiency but risks overfitting to synthetic patterns. 
Multi-task tuning (ROUTE~\cite{qin2025route}) boosts multi-step reasoning but complicates training convergence. 
While current methods focus on isolated improvements, the field lacks holistic frameworks to combine architectural technical contributions. 
Future FT methods need choreographed pipelines that sequentially optimize pre-training objectives, data diversity strategies and task-specific adaptations.
This orchestration will be crucial for open-source models to achieve the multifaceted capabilities of proprietary LLMs in text-to-SQL.

\subsubsection{Comparing ICL and FT paradigm}
The paradigms present asymmetric value propositions. 
ICL methods leverage proprietary LLMs (e.g., GPT-4~\cite{achiam2023gpt4}) for rapid adaptation but face constraints in controllability and domain customization. 
FT methods empower open-source models (e.g., CodeLLaMA~\cite{roziere2023codellama}) with targeted SQL capabilities through fine-tuning but require substantial engineering for data curation and training optimization. 
Importantly, FT methods enhance foundational model competence: CodeS's~\cite{li2024codes} incremental pre-training boosts question and schema understanding, while SQL-LLaMA's~\cite{wang2024macsql} multi-agent tuning supports complex compositional reasoning. 
Nonetheless, ICL methods' prompt-based flexibility remain unmatched for addressing emerging database environments without retraining. 
The future landscape will likely reveal a symbiotic relationship: FT develops specialized models with general SQL grounding, while ICL offers lightweight adaptation for downstream application. 
This division mirrors trends in a retrieval-augmented generation, where FT builds core capabilities, and ICL manages dynamic context integration.

\section{Expectations}
Despite the significant advancements made in LLM-based text-to-SQL research, there are still several challenges that need to be addressed. In this section, we discuss the remaining challenges that we expect to overcome in future work.
\subsection{Robustness in Real-world Applications}
The text-to-SQL implemented by LLMs is expected to perform generalization and robustness across complex scenarios in real-world applications. 
Despite recent advances having made substantial progress in robustness-specific datasets~\cite{pi2022ADVETA,deng2021Spider-Realistic}, its performance still falls short of practical application~\cite{li2023BIRD}. 
There are still challenges that are expected to be overcome in future studies. From the user aspect, there is a phenomenon that the \textbf{user is not always a clear question proposer}, which means the user questions might not have the exact database value and also can be varied from the standard datasets, the synonyms, typos, and vague expressions could be included~\cite{gan2021Spider-SYN}.
For instance, the models are trained on clear indicative questions with concrete expressions in the fine-tuning paradigm.
Since the model has not learned the mapping of realistic questions to the corresponding database, this leads to a knowledge gap when applied to real-world scenarios~\cite{li2023BIRD}.
As reported in the corresponding evaluations of the dataset with synonym and incomplete instruction~\cite{rajkumar2022evaluating,tai2023exploring}, the SQL queries generated by ChatGPT contain around 40\% incorrect execution, which is 10\% lower than the original evaluation~\cite{tai2023exploring}.
For the evaluation in real-world enterprise tasks~\cite{lei2025spider2}, current state-of-the-art models solve only about 21.3\% of these tasks, far below their performance on earlier benchmarks~\cite{yu2018Spider,li2023BIRD}.
Simultaneously, the \textbf{fine-tuning with local text-to-SQL datasets may contain non-standardized samples and labels.}
As an example, the name of the table or column is not always an accurate representation of its content, which yields an inconsistency within the training data construction and may lead to a semantic gap between the database schema and the user question. 
To address this challenge, aligning the LLMs with intention bias and designing the training strategy towards noisy scenarios will benefit the recent advances. 
At the same time, \textbf{the data size in real-world applications is relatively smaller than the research-oriented benchmark}. 
Since extending a large amount of the data by human annotation incurs high labor costs, designing data-augmentation methods to obtain more question-SQL pairs will support the LLM in data scarcity.
Besides that, the adaptation study of fine-tuned open-source LLM to local small-size datasets can be potentially beneficial.
Furthermore, \textbf{the extensions on {multi-lingual}~\cite{min2019CSpider,tuan2020pilot} and multi-modal scenarios}~\cite{song2024speech} should be studied in future research, which will benefit more language groups and help build more general database interfaces. 

\subsection{Computational Efficiency}
The computational efficiency is determined by the inference speed and the cost of computational resources, which is worth considering in both application and research work~\cite{kou2024cllms,zhang2023actsql}.
With the increasing complexity of databases in up-to-date text-to-SQL benchmarks~\cite{pourreza2023evaluating,li2023BIRD,lei2025spider2}, databases will carry more information (including more tables and columns), and the token length of the database schema will correspondingly increase, raising a series of challenges.
Dealing with an ultra-complex database, taking the corresponding schema as input may encounter the challenge that \textbf{the cost of calling proprietary LLMs will significantly increase, potentially exceeding the model's maximum token length}, especially with the implementation of open-source models that have shorter context lengths.
Meanwhile, another obvious challenge is that most works \textbf{use the full schema as model input, which introduces significant redundancy}~\cite{wang2024macsql}.
Providing LLMs with a precise question-related filtered schema directly from the user end to reduce cost and redundancy is a potential solution to improve computational efficiency~\cite{dong2023c3}. 
Designing an accurate method for schema filtering remains a future direction.
Although the in-context learning paradigm achieves promising accuracy, as a computational efficiency concern, the well-designed methods with the multi-stage framework or extended context \textbf{increasing the number of API calls to enhance performance has simultaneously led to a substantial rise in costs}~\cite{pourreza2023dinsql}.
Recent agentic reasoning approaches, relying on iterative multi-step actions, typically multiply API calls severalfold. While enabling complex task handling and improved robustness, optimizing action planning to reduce redundant interactions across diverse database scenarios remains an important direction for future research.
As reported in the related study~\cite{zhang2023actsql}, a trade-off between performance and computational efficiency should be considered carefully, and designing a comparable (even better) in-context learning method with less API cost will be a practical implementation and is still under exploration.
Compared to PLM-based methods, \textbf{the inference speed of LLM-based methods is observably slower}~\cite{yang2024harnessing,li2023resdsql}. 
Accelerating inference by shortening the input length and reducing the number of stages in implementation would be intuitive for the in-context learning paradigm.
For local LLMs, from a starting point~\cite{kou2024cllms}, more speedup strategies can be studied in enhancing the model's architecture in future exploration. 

\subsection{Data Privacy and Interpretability}
As a part of the LLMs' study, LLM-based text-to-SQL also faces some general challenges present in LLM research~\cite{yan2024protecting,zhang2023siren,singh2024rethinking}.
Potential improvements from the text-to-SQL perspective are also expected to be seen in these challenges, thereby extensively benefiting the study of LLMs.
As previously discussed in Sec.~\ref{sec:icl}, the in-context learning paradigm predominates the number and performance in recent studies, with the majority of work using proprietary models for implementation~\cite{pourreza2023dinsql,gao2023dailsql}.
A straightforward challenge is proposed regarding data privacy, as \textbf{calling proprietary APIs to handle local databases with confidentiality can pose a risk of data leakage}. 
Using a local fine-tuning paradigm can partially address this issue. 
Still, the current performance of vanilla fine-tuning is not ideal~\cite{gao2023dailsql}, and advanced fine-tuning framework potentially relies on proprietary LLMs for data augmentation~\cite{li2024codes}.
Based on the current status, more tailored frameworks in the local fine-tuning paradigm for text-to-SQL deserve widespread attention.
Overall, the development of deep learning continually faces challenges regarding interpretability~\cite{singh2024rethinking,dai2022knowledge}. 
As a long-standing challenge, considerable work has already been studied to address this issue~\cite{zhang2024comprehensive,meng2023massediting}.
However, in text-to-SQL research, \textbf{the interpretability of LLM-based implementation is still not being discussed}, whether in the in-context learning or fine-tuning paradigm.
The approaches with a decomposition phase explain the text-to-SQL implementation process from the perspective of step-by-step generation~\cite{pourreza2023dinsql,tai2023exploring}. 
Building on this, combining advanced study in interpretability~\cite{meng2022locating,zheng2023edit} to enhance text-to-SQL interpreting the model architecture from the database knowledge aspect remains a future direction.

\subsection{Extensions}
As a sub-field of LLMs and natural language understanding research, many studies in these fields have been adopted for text-to-SQL tasks, advancing its development~\cite{wei2022chain,wang2023selfconsistency}. 
However, text-to-SQL research can also be extended to the larger scope studies of these fields at meanwhile.
For instance, SQL generation is a part of code generation. 
The well-designed approaches in code generation also obtain promising performance in text-to-SQL~\cite{ni2023lever,chen2024selfdebugging}, performing generalization across various programming languages.
\textbf{The potential extension of some tailored text-to-SQL frameworks to NL-to-code studies can also be discussed}.
For instance, frameworks integrating execution output in NL-to-code can also achieve solid performance in SQL generation~\cite{pourreza2023dinsql}.

An attempt to extend execution-aware approaches in text-to-SQL with other advancing modules~\cite{hong2024knowledgetosql,dong2023c3} for code generation is worth discussing.
From another perspective, we previously discussed that text-to-SQL can enhance LLM-based question-answering (QA) by providing factual information.
The database can store relational knowledge as structural information, and \textbf{the structure-based QA can potentially benefit from text-to-SQL} (e.g., knowledge-based question-answering, KBQA~\cite{luo2023chatkbqa,li2024flexkbqa}).
Construct the factual knowledge with database structure, and then incorporate the text-to-SQL system to achieve information retrieval, which can potentially assist further QA with more accurate factual knowledge~\cite{xiong2024interactive}. 
More extensions of text-to-SQL studies are expected in future work.

\bibliographystyle{IEEEtran}
\bibliography{reference}

\begin{IEEEbiography}[{\includegraphics[width=1in,height=1.25in,clip,keepaspectratio]{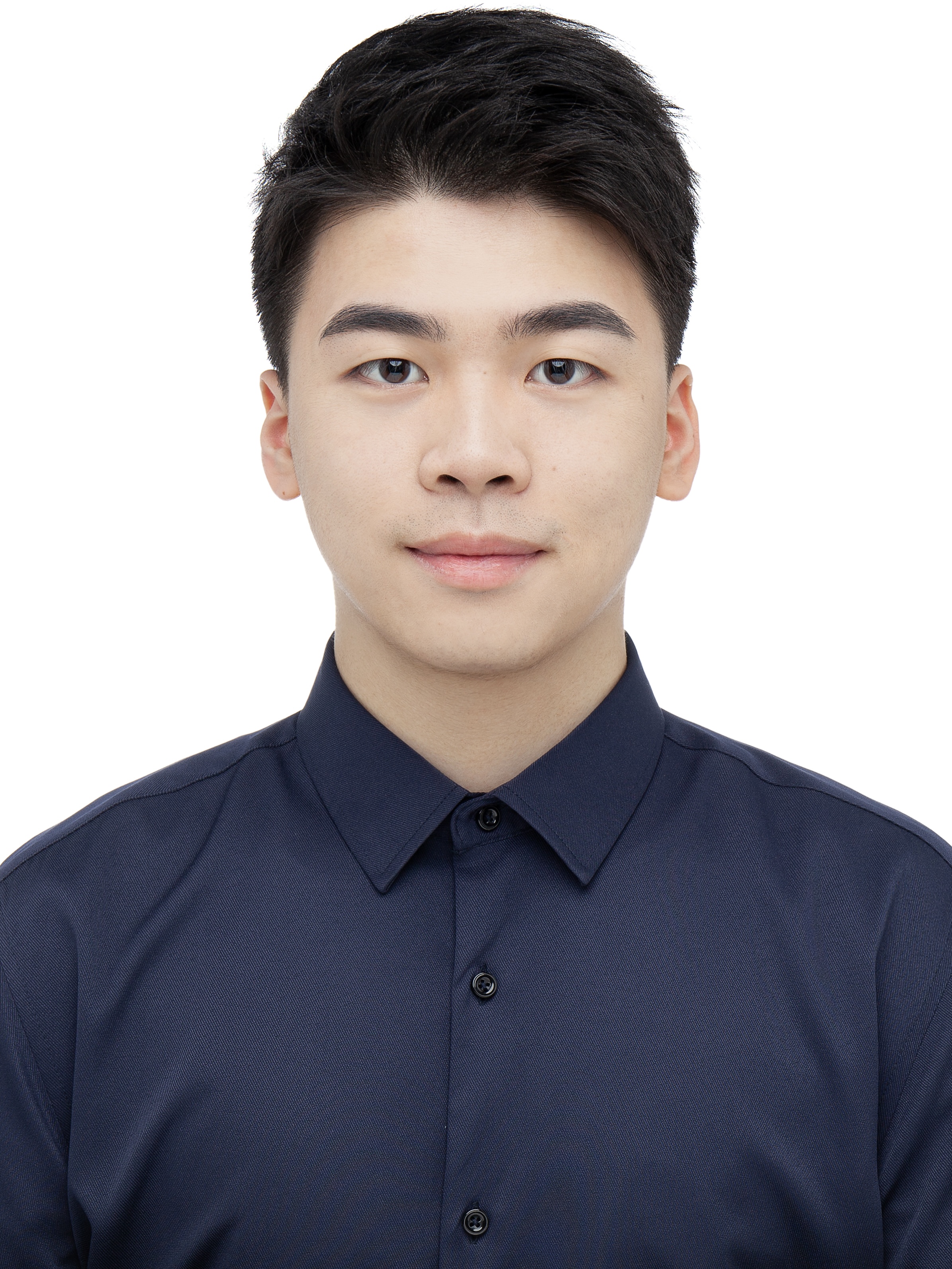}}]{Zijin Hong} is a first-year Ph.D. student in the DEEP Lab, Department of Computing, The Hong Kong Polytechnic University, under the supervision of Prof. Xiao Huang. He holds a dual bachelor’s degree in Applied Mathematics with Mathematics from Jinan University and the University of Birmingham. His research interests include machine learning, large language models, and their applications, particularly LLM-based text-to-SQL. He has published papers and served as a reviewer for venues such as ACL, EMNLP, NAACL, AAAI, ICLR, and IEEE TKDE.
\end{IEEEbiography}

\begin{IEEEbiography}[{\includegraphics[width=1in,height=1.25in,clip,keepaspectratio]{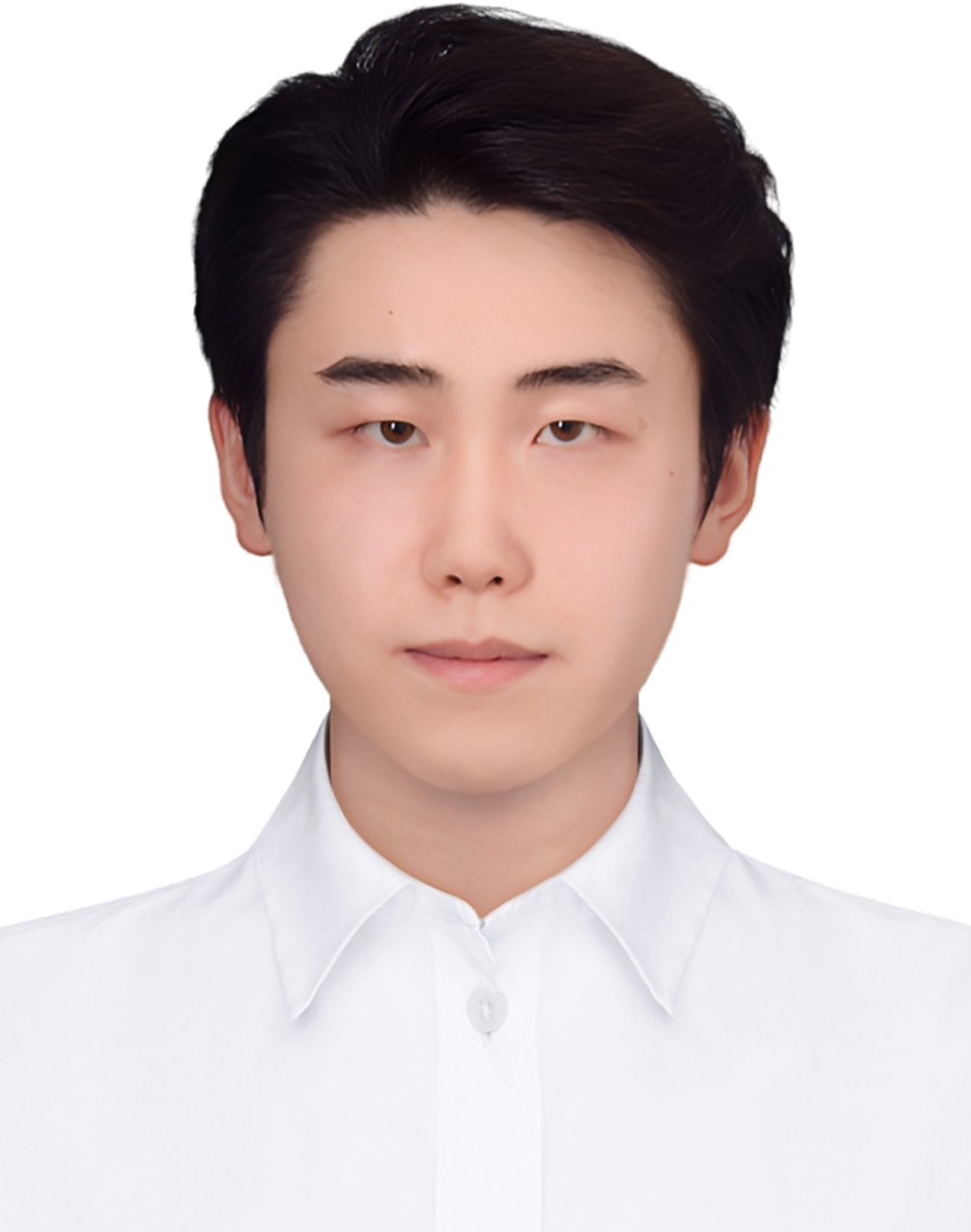}}]{Zheng Yuan} is a second-year Ph.D. student at the Department of Computing, The Hong Kong Polytechnic University. He received a bachelor degree in Computer Science and Technology from Shenzhen University in 2021. He is currently a group member in the DEEP Lab supervised by Prof. Xiao Huang at the Hong Kong Polytechnic University. His research interests focus on large language models and their applications in data-driven scenarios, including retrieval-augmented generation, text-to-SQL, and recommender systems.
\end{IEEEbiography}

\begin{IEEEbiography}[{\includegraphics[width=1in,height=1.25in,clip,keepaspectratio]{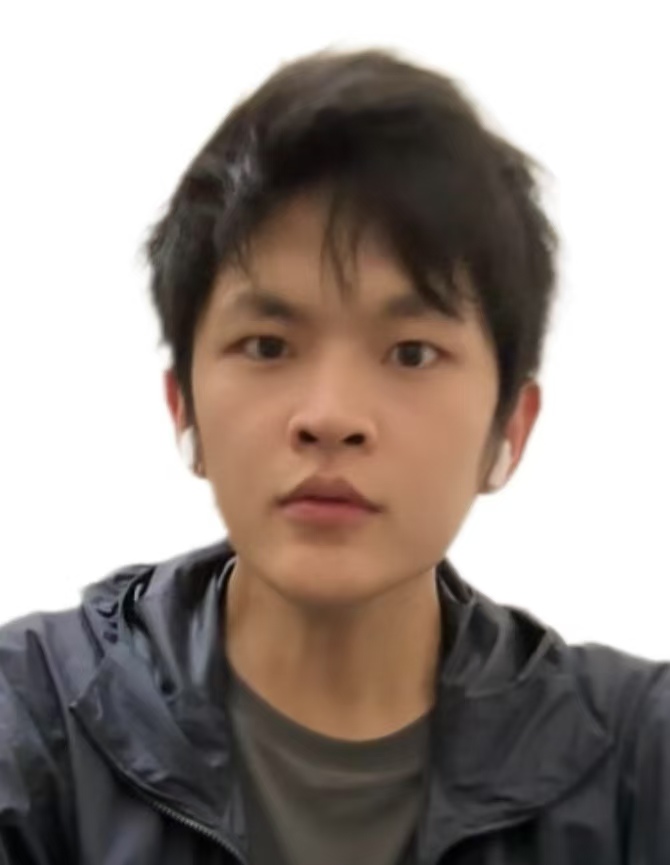}}]{Qinggang Zhang} is currently a Postdoctoral Fellow at the Deep Lab, Department of Computing, The Hong Kong Polytechnic University, Hong Kong SAR. He received his Ph.D. and B.Eng. degrees from The Hong Kong Polytechnic University and Northwestern Polytechnical University, respectively. His research interests include KGs, LLMs, RAG, and Text-to-SQL. He has published over 20 papers while serving as a reviewer for NeurIPS, ICML, ICLR,  KDD, IEEE TKDE, and IEEE TPAMI.
\end{IEEEbiography}

\begin{IEEEbiography}[{\includegraphics[width=1in,height=1.25in,clip,keepaspectratio]{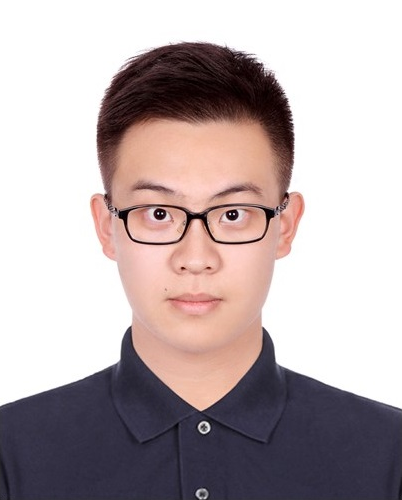}}]{Hao Chen} is an assistant professor at the Faculty of Data Science, City University of Macau. Before that, he worked as a postdoctoral research fellow at The Hong Kong Polytechnic University. He received his Ph.D degree from Beihang University, China in 2022. He has published over 40 papers, such as TKDE, NeurIPS, ACL, SIGKDD, WWW, and SIGIR, etc. He received the Best Paper Award Honorable Mention at SIGIR 2023. His research interests include recommender systems, graph neural networks, and large language models.
\end{IEEEbiography}

\begin{IEEEbiography}[{\includegraphics[width=1in,height=1.25in,clip,keepaspectratio]{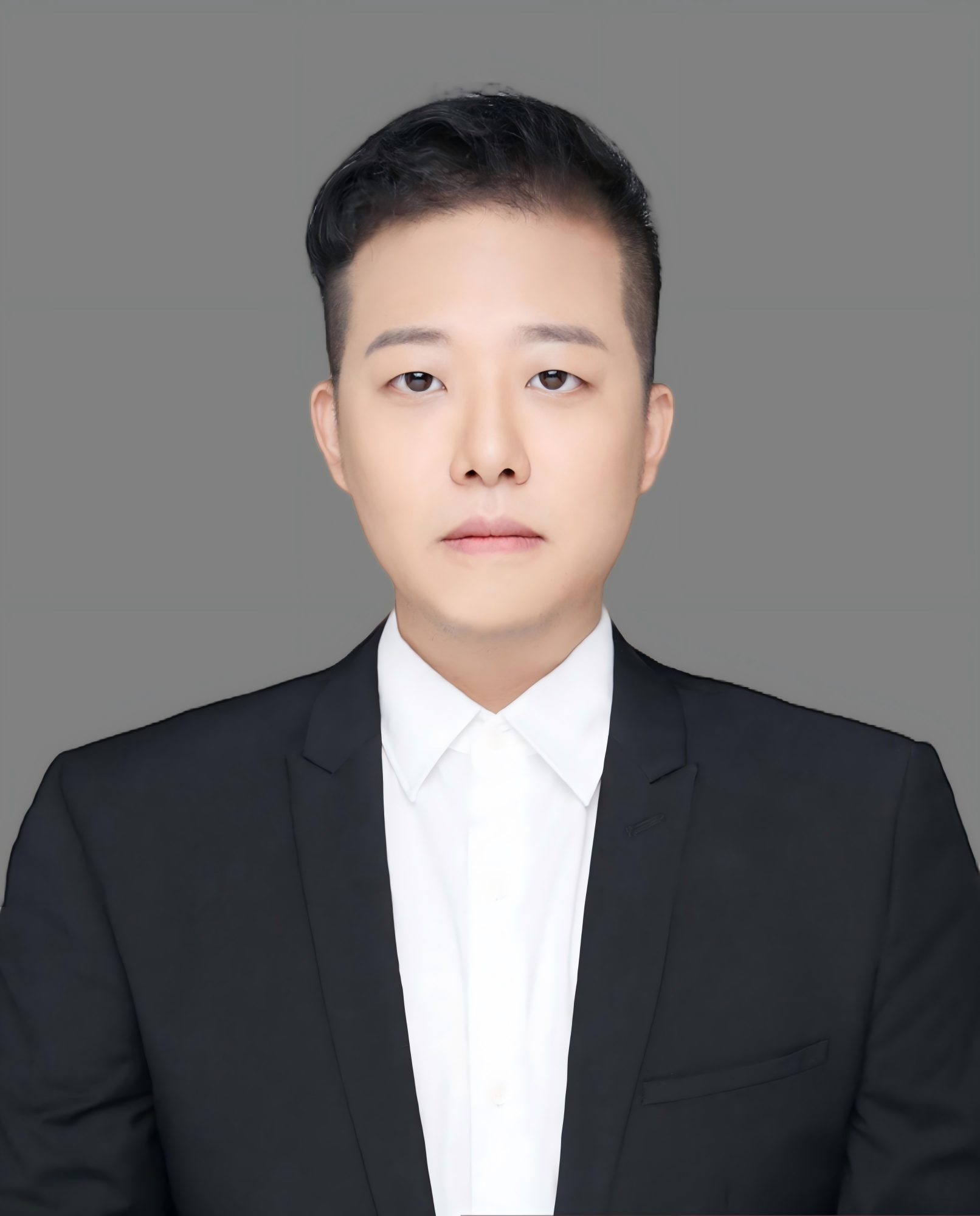}}]{Junnan Dong} is the final-year Ph.D. candidate in the department of computing, Hong Kong Polytechnic University. He focuses on Large Language Models, Knowledge Graphs and Graph Retrieval Augmented Generation for advanced and knowledgeable domain-specific reasoning. He has published more than ten papers on top-tier conferences, including NeurIPS, ACL, WWW, KDD, SIGIR and WSDM. He has been continuously serving as a program committee member for NeurIPS, ICML, ICLR, ACL and WWW.
\end{IEEEbiography}

\begin{IEEEbiography}[{\includegraphics[width=1in,height=1.25in,clip,keepaspectratio]{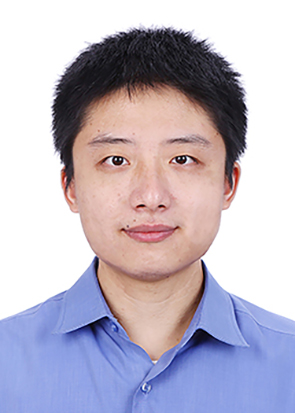}}]{Feiran Huang} received the Ph.D. degree from the School of Computer Science and Engineering, Beihang University, Beijing, China, in 2018. He is currently a professor with the College of Information Science and Technology, Jinan University. He has over 100 publications appearing in top conferences and journals, such as SIGIR, KDD, WWW, MM, TIP, IJCV, and TKDE. He received 5 best paper awards, such as the Best Paper Award Honourable Mention in SIGIR 2023 and Best Paper Award Runner Up in PAKDD 2023. He holds over 50 US, Chinese, and international granted patents. He is an editorial board member of IEEE Transactions on Affective Computing, ACM Transactions on Recommender Systems, etc. His research interests include recommender systems, social networks, sentiment analysis, and large language models.
\end{IEEEbiography}

\begin{IEEEbiography}[{\includegraphics[width=1in,height=1.25in,clip,keepaspectratio]{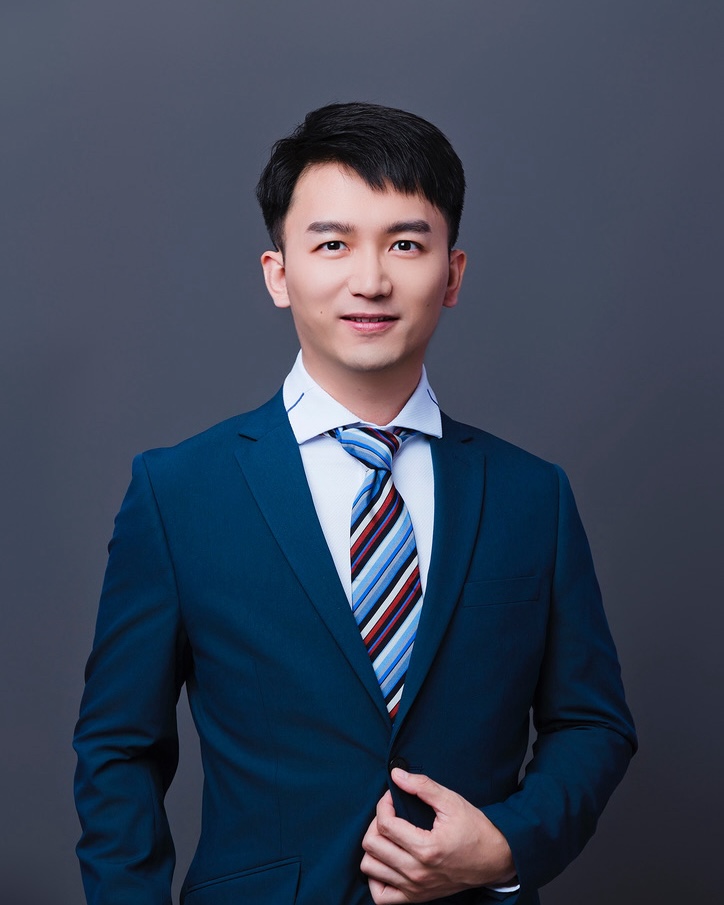}}]{Xiao Huang} is an Assistant Professor in the Department of Computing at The Hong Kong Polytechnic University. He earned his Ph.D. in Computer Engineering from Texas A\&M University in 2020, an M.S. in Electrical Engineering from the Illinois Institute of Technology in 2015, and a B.S. in Engineering from Shanghai Jiao Tong University in 2012. His scholarly contributions are highly regarded within the academic community, amassing over 4,500 citations. 
He received the Best Paper Award Honorable Mention at SIGIR 2023. He has successfully led or completed seven research projects as Principal Investigator.
\end{IEEEbiography}

\end{document}